\documentclass[letterpaper, 10 pt, conference]{ieeeconf}  

\IEEEoverridecommandlockouts                              

\usepackage[top=19.1mm, bottom=19.1mm, left=19.1mm, right=19.1mm]{geometry}
\usepackage{lipsum}
\usepackage{colortbl}
\usepackage{url}
\usepackage{graphicx}
\usepackage{amsfonts}
\usepackage{soul} 
\usepackage{color, xcolor} 
\usepackage{amsmath}
\usepackage{amssymb}
\usepackage{multirow}
\usepackage{array}
\usepackage[linesnumbered,ruled,vlined]{algorithm2e}
\usepackage{booktabs}
\usepackage[noadjust]{cite}
\usepackage[colorlinks,allcolors=black]{hyperref}
\usepackage{color}
\usepackage[caption=false, font=footnotesize]{subfig}

\usepackage{threeparttable}
\usepackage{arydshln}

\title{\LARGE \bf
MAER-Nav: Bidirectional Motion Learning Through Mirror-Augmented Experience Replay for Robot Navigation
}

\author{Shanze Wang$^{1, 2}$, Mingao Tan$^{2}$, Zhibo Yang$^{3}$, Biao Huang$^{2,4}$, Xiaoyu Shen$^{2}$, \\Hailong Huang$^{1}$ and Wei Zhang$^{2}$
\thanks{This work has been submitted to the IEEE for possible publication. Copyright may be transferred without notice, after which this version may no longer be accessible.}
\thanks{*This work is supported by 2035 Key Research and Development Program of Ningbo City under Grant No.2024Z127. (\textit{Corresponding author: Wei Zhang.})}
\thanks{$^{1}$Shanze Wang and Hailong Huang are with the Department of Aeronautical and Aviation Engineering, Hong Kong Polytechnic University,
        {\tt\small shanze.wang@connect.polyu.hk; hailong.huang@polyu.edu.hk}}%
\thanks{$^{2} $Shanze Wang, Mingao Tan, Biao Huang, Xiaoyu Shen and Wei Zhang are with the Ningbo Key Laboratory of Spatial Intelligence and Digital Derivative, Institute of Digital Twin, Eastern Institute of Technology, Ningbo, China,
        {\tt\small szwang@eitech.edu.cn; mtan@eitech.edu.cn; 210810114@stu.hit.edu.cn; xyshen@eitech.edu.cn; zhw@eitech.edu.cn}}%
\thanks{$^{3}$Zhibo Yang is with the Department of Mechanical Engineering, National University of Singapore,
        {\tt\small zhibo.yang@u.nus.edu}}%
\thanks{$^{4}$Biao Huang is with the School of Science, Harbin Institute of Technology, Shenzhen,{\tt\small 210810114@stu.hit.edu.cn}}%
}

\begin{document}

\maketitle
\thispagestyle{empty}
\pagestyle{empty}

\begin{abstract}
Deep Reinforcement Learning (DRL) based navigation methods have demonstrated promising results for mobile robots, but suffer from limited action flexibility in confined spaces. Conventional DRL approaches predominantly learn forward-motion policies, causing robots to become trapped in complex environments where backward maneuvers are necessary for recovery. This paper presents MAER-Nav (Mirror-Augmented Experience Replay for Robot Navigation), a novel framework that enables bidirectional motion learning without requiring explicit failure-driven hindsight experience replay or reward function modifications. Our approach integrates a mirror-augmented experience replay mechanism with curriculum learning to generate synthetic backward navigation experiences from successful trajectories. Experimental results in both simulation and real-world environments demonstrate that MAER-Nav significantly outperforms state-of-the-art methods while maintaining strong forward navigation capabilities. The framework effectively bridges the gap between the comprehensive action space utilization of traditional planning methods and the environmental adaptability of learning-based approaches, enabling robust navigation in scenarios where conventional DRL methods consistently fail.

\end{abstract}

\section{Introduction}

Mobile robot navigation has become increasingly important across various applications, from warehouse automation and healthcare services to retail environments and last-mile delivery \cite{cognominalEvolvingFieldAutonomous2021}, \cite{srinivasAutonomousRobotdrivenDeliveries2022}, \cite{hossainAutonomousDeliveryRobots2023}. Deep Reinforcement Learning (DRL) functions as an effective computational framework for autonomous navigation by facilitating the acquisition of optimal control policies through direct environmental interaction and reward signal optimization without manual behavior design. Current research primarily focuses on navigation in highly constrained environments \cite{liu_lifelong_2021} and scenarios involving dynamic obstacles or social interactions with pedestrians  \cite{xie2023drl}, \cite{mavrogiannisCoreChallengesSocial2023}. The complexity of these environments, combined with the need for real-time decision-making, makes safety and flexibility essential considerations in autonomous navigation systems. Consequently, path planning has become a critical component that directly influences the robot's ability to operate efficiently and safely in these challenging scenarios.

\begin{figure}[t]
	\centering
	\includegraphics[width=\linewidth]{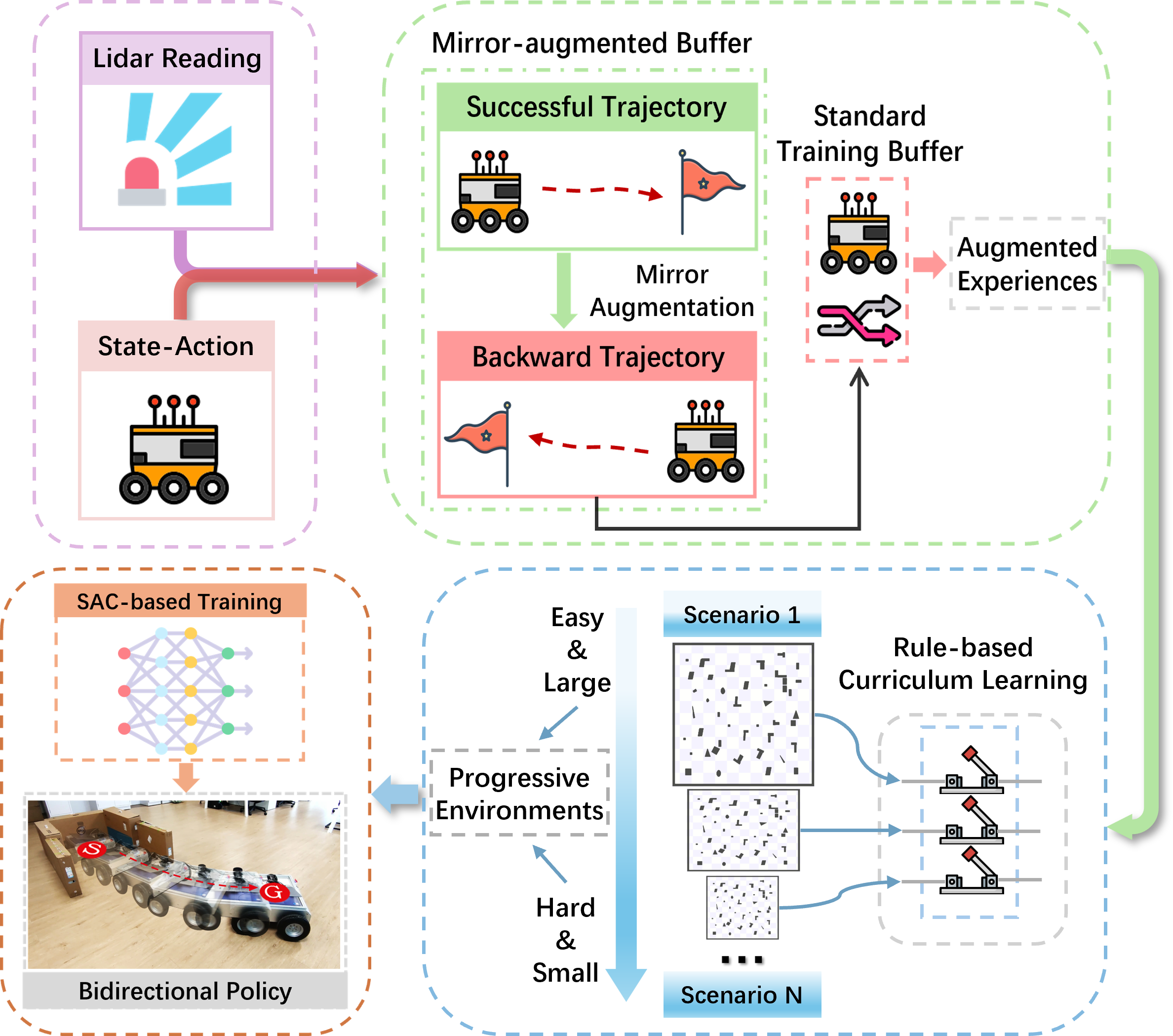}
	\caption{Proposed MAER-Nav structure. By mirroring the states and actions of successful episodes in the mirror-augmented buffer, we aim to enable the local planner trained by the DRL-based method to learn the bidirectional capability.}
	\label{whole structure fig1}
\end{figure}

However, DRL-based navigation local planners face a significant challenge regarding action flexibility. The behavior of DRL methods heavily depends on training experiences, and in conventional navigation training, successful trajectories predominantly consist of forward movements. This bias leads to models developing a strong preference for forward motion. 
Empirical studies [9], [10], [11], [12] have demonstrated that DRL-trained robots possess limited capacity for executing bidirectional movements effectively, with evidence showing they frequently encounter difficulties in complex scenarios where reverse movement would represent the optimal solution [13].
In contrast, traditional planning methods such as Dynamic Window Approach (DWA) [7] and Timed Elastic Band (TEB) [8] consider the complete velocity space during trajectory search, including both forward and backward motions. These methods utilize explicit mathematical models to evaluate all possible movement options without directional bias. 

To address this challenge, we propose MAER-Nav (Mirror-Augmented Experience Replay for Robot Navigation), a novel DRL-based framework specifically designed to solve action flexibility limitations in robot navigation. Unlike classical hindsight experience replay (HER) \cite{andrychowicz2017hindsight} methods that rely on relabeling failed trajectories, our approach introduces a mirror-augmented experience replay (MAER) mechanism that creates synthetic backward navigation experiences directly from successful trajectories. This removes the need for explicit failure-driven hindsight experience replay while preserving the reliability of original successful episodes. Additionally, we employ an adaptive curriculum learning strategy to dynamically adjust training scenarios based on performance metrics, gradually exposing the robot to increasingly complex environments.
The key contributions of this work are as follows:
\begin{itemize}
    \item A \textbf{Mirror-Augmented Experience Replay (MAER)} mechanism that develops bidirectional navigation capabilities from successful trajectories, enabling comprehensive motion learning independent of explicit failure-driven hindsight experience replay.
    \item A \textbf{Bidirectional Action Recovery} mechanism that simulates recovery behaviors through synthetic trajectory generation, addressing the action bias in conventional DRL training.
    \item Extensive simulation and real-world experiments demonstrating significant improvements in navigation capabilities, particularly in the robot's ability to execute effective backward maneuvers when compared to conventional DRL-based methods.
\end{itemize}

\section{Related Work}

\subsection{Traditional Robot Navigation Methods}
Traditional local planners form the foundation of autonomous navigation, with DWA \cite{foxDynamicWindowApproach1997a}, E-Band \cite{quinlanElasticBandsConnecting1993a}, and TEB \cite{rosmann2012trajectory} being the most widely implemented methods. DWA selects optimal velocity commands within a dynamic window to ensure safety and efficiency. E-Band utilizes elastic band theory to refine paths through contraction and repulsion mechanisms, while TEB enhances this approach by integrating time optimization with velocity and acceleration constraints, making it effective in dynamic environments.
Despite their proven effectiveness in standard scenarios, these traditional planners exhibit notable limitations in complex settings. DWA predominantly favors forward motion and rarely executes backward maneuvers unless forward movement is impossible. E-Band becomes unstable during sudden environmental changes, resulting in oscillatory path adjustments, while TEB achieves smoother trajectories but at significant computational expense. These approaches collectively struggle with navigational performance in complex scenarios and fail to meet real-time requirements as environmental complexity increases, thus highlighting the need for DRL-based methods that offer more robust navigation solutions.

\subsection{DRL-Based Robot Navigation}

The performance of traditional algorithms deteriorates significantly as environmental complexity increases. In contrast, DRL-based algorithms excel in processing high-dimensional, complex inputs and continuous action spaces, making them well-suited for path planning. Hu et al. \cite{huDeepReinforcementLearningBased2025} address local optima challenges by designing reward rules for exploration and improving the SAC algorithm \cite{haarnoja2018soft} with an adaptive temperature parameter. Through subgoal selection and target-directed action representation, a hierarchical DRL framework enhances navigation generalization ability, safety, and sim-to-real transferability \cite{zhuHierarchicalDeepReinforcement2023}. To address the challenge of long-range DRL navigation under crowded conditions, Jing et al. \cite{jingTwoStageReinforcementLearning2024a} propose a two-stage DRL method, which trains the agent to generate safe subgoals and refine planning for efficient and safe movement. In addition, experience replay improves learning by storing past experiences, mixing old and new data to break patterns, and reusing rare experiences for better training \cite{linSelfimprovingReactiveAgents1992}. Instead of uniformly sampling from replay memory, Schaul et al. \cite{schaulPrioritizedExperienceReplay2016} propose a framework for prioritizing experience, enabling more frequent replay of important transitions and subsequently improving learning efficiency.

Although DRL-based local planners have demonstrated effectiveness in autonomous navigation, they generally lack explicit reverse movement capabilities. For common circular robots, where the drive center typically coincides with the geometric center, backward motion can be substituted by in-place rotation followed by forward movement, which leads most research in this area to disregard bidirectional movement capabilities \cite{mo_adobeindoornav_2018}. However, for other types of non-omnidirectional robots, this capability is crucial. Furthermore, some studies deliberately restrict backward action or impose penalties for reversing, thereby overlooking situations where such maneuvers are necessary \cite{majid2024challenging}, \cite{zhuDeepReinforcementLearning2021}. 


\section{Preliminaries}
\subsection{Action Flexibility Constraints in DRL-Based Training}

\begin{figure*}[t]
	\centering
	\includegraphics[width=0.95\linewidth]{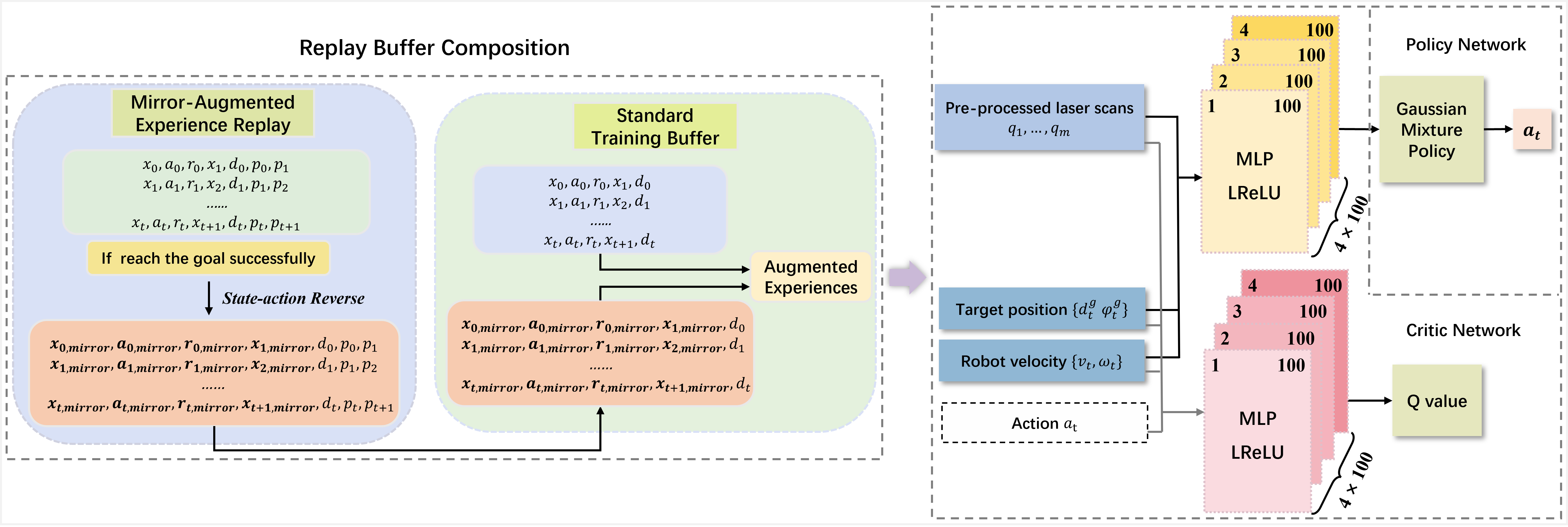}
	\caption{The overall network structure of MAER-Nav.}
	\label{network}
\end{figure*}

Current DRL-based navigation methods demonstrate limited action flexibility, particularly regarding backward motion capabilities. This constraint stems from several factors in existing research \cite{taheri_deep_2024}. As existing navigation frameworks rely heavily on forward-facing sensors and the absence of backward sensors, researchers have to restrict the action space to forward motion and rotation \cite{jiang_itd3-cln_2022}, \cite{chen_deep_2021}, \cite{jestel_obtaining_2021}, \cite{yao_crowd-aware_2021}. The navigation framework presented in \cite{kastner_arena-rosnav_2021} applies penalties to backward movement in the reward function, reflecting that such movement is undesirable in navigation environments. Even when systems theoretically permit backward motion via negative linear velocities, training protocols often restrict these actions \cite{zhu_hierarchical_2023}, \cite{zhu_target-driven_2016}.
Systems that do incorporate backward motion typically implement constraints that favor combining forward motion with rotation, limiting backward movement for emergency scenarios only \cite{wu_reinforcement_2021}. This approach significantly reduces maneuverability in complex environments. Additionally, when backward motion is permitted, its effectiveness is often compromised by fixed step sizes in both forward distance and rotation angles, resulting in discontinuous action transitions \cite{hu_sim--real_2021}. These limitations are intensified by DRL's experience-dependent nature—since successful training trajectories predominantly feature forward movements, models develop strong forward motion biases regardless of their theoretical action space \cite{han_improved_2023}.

This pervasive forward bias in current DRL-based navigation systems represents a significant limitation in real-world applications, particularly in confined spaces where backward maneuvers are essential for successful navigation. Our work directly addresses these constraints by developing methods for effective backward motion learning using only 2D LiDAR input, without modifying the reward function or adding more sensors.

\subsection{Problem Formulation}
This study addresses the action flexibility limitations of DRL-based navigation methods. Conventional DRL approaches implement forward-only policies with standard experience buffers, which often causes robots to become trapped in confined spaces. Our proposed MAER-Nav method introduces a bidirectional policy framework with an augmented experience replay buffer that enables effective backward motion capabilities. We formulate this problem for a differential-drive rectangular robot equipped with 2D LiDAR that must reach designated goals while maintaining full action space utilization.

At time step $t$, the DRL agent’s observation is represented as $x_t = \{L_t, x_t^g, v_t, \omega_t\}$, where  $L_t$ denotes the min-pooled LiDAR readings.  $x_t^g$ represents the target position, and $v_t$ and $\omega_t$ are the robot’s current linear and angular velocities, respectively. 
Upon receiving the current observation $x_t$, the agent determines and implements an action $a_t$ (velocity command) in accordance with policy $\pi$. Following the implementation of $a_t$, the system updates to observation $x_{t+1}$ incorporating new sensory data,and provides the agent with a reward $r_t$ as follows:
\begin{equation}
\begin{aligned}
r_t = 
\begin{cases}
r_{s}, & \text{if success,}\\
r_{c}, & \text{if crash,}\\
c_{1}\left(d_{t}^{g}-d_{t+1}^{g} \right), & \text{otherwise}.
\end{cases}
\end{aligned}
\end{equation}
where $c_1$ is a coefficient for scaling. The reward mechanism incorporates three distinct elements: a positive term $r_s$ that rewards task completion, a negative penalty $r_c$ applied when collisions occur, and a minor continuous reward designed to encourage directional movement toward the objective. The goal of this work is to find an optimal policy $\pi^\ast$ that maximizes the discounted total reward $G_t = \sum_{\tau=t}^{T} \gamma^{\tau-t} r_\tau$, where $\gamma \in \left[0,1\right)$ is the discount factor. In this paper, the parameters are set as $c_1=2$, $r_s=10$, $r_c = -10$, and $\lambda=0.99$.

\section{Approach}

Fig. \ref{whole structure fig1} presents the overall architecture of the proposed MAER-Nav method. The framework consists of a bidirectional action recovery mechanism, a dual-buffer experience storage system, and curriculum learning that gradually increases environmental complexity. This structure enables robots to acquire bidirectional navigation capabilities without requiring explicit failure-driven hindsight experience replay or additional sensors. The following subsections detail each component of the MAER-Nav framework.

\subsection{Trajectory Synthesis via Bidirectional Action Recovery}

The core innovation of MAER-Nav lies in its ability to transform successful trajectories into bidirectional motion experiences. Unlike traditional DRL approaches that only learn from forward motion patterns, our method introduces a novel trajectory synthesis mechanism that enables learning from both forward and backward motion patterns while maintaining physical consistency.

For a successful trajectory $\tau = \{(x_t, a_t, r_t, x_{t+1})\}_{t=0}^{T}$, where $x_t$ represents the agent's observation (which serves as the state) at time $t$, $a_t$ the action taken, $r_t$ the reward received, and $x_{t+1}$ the resulting state, our bidirectional action recovery mechanism operates by analyzing the inherent symmetry in robot navigation tasks. The key insight is that for any successful forward trajectory to a goal, there exists a corresponding valid backward trajectory that could navigate from that goal to the starting position.
This symmetry is particularly important for differential-drive robots, where the motion capabilities are fundamentally bidirectional but traditional learning approaches fail to utilize backward motion effectively. Our approach formalizes this bidirectional capability through a state-action transformation framework:

\begin{equation}
\mathcal{T}: (x_t, a_t) \rightarrow (x_{t,\mathrm{\text{mirror}}}, a_{t,\mathrm{\text{mirror}}})
\end{equation}

The transformation $\mathcal{T}$ preserves the physical constraints of the robot while enabling reverse motion learning. For a differential-drive robot, this transformation should consider:

\begin{equation}
x_{t,\mathrm{\text{mirror}}} = \{L_t, {x_{t,\mathrm{\text{mirror}}}^g}, v_t, \omega_t\}
\end{equation}
where ${x_{t,\mathrm{\text{mirror}}}^g}$ represents the transformed goal coordinates in the robot's reference frame. The critical aspect of this transformation is the directional inversion implemented through ${x_{t,\mathrm{\text{mirror}}}^g}$. In our bidirectional action recovery mechanism, the original goal position is treated as the new start point, while the original start position becomes the new target goal.

Simply but effectively, the action transformation is achieved through direct negation:

\begin{equation}
a_{t,\mathrm{\text{mirror}}} = -a_t
\end{equation}

By maintaining the spatial relationship between the robot and its environment while inverting the direction of motion, the robot can effectively utilize experiential data for comprehensive navigation learning without requiring separate training processes for forward and backward motion.

\subsection{Implementation of Mirror-Augmented Experience Storage}
We implement a dual-buffer experience replay architecture for effective bidirectional motion learning. Our approach utilizes two complementary buffers: a standard experience replay buffer $\mathcal{B}$ as primary transition repository and a specialized mirror-augmented buffer $\mathcal{B}_{\mathrm{\text{mirror}}}$ for specific-episode storage.
The standard training buffer $\mathcal{B}$ maintains the conventional experience tuples:
\begin{equation}
\mathcal{B} = \{(x_t, a_t, r_t, x_{t+1}, d_t)\}_{t=0}^{T}
\end{equation}
where $x_t$ represents the observation including LiDAR readings and robot state, $a_t$ is the action, $r_t$ is the reward, $d_t$ is the terminal flag.
To enable bidirectional motion learning, we introduce a specialized mirror-augmented buffer $\mathcal{B}_{\text{mirror}}$ that captures additional pose information for each episode:

\begin{equation}
\mathcal{B}_{\mathrm{\text{mirror}}} = \{(x_t, a_t, r_t, x_{t+1}, d_t, p_t, p_{t+1})\}_{t=0}^{T}
\end{equation}
where $p_t = (p^x_t, p^y_t, \theta_t)$ captures the robot pose in the global frame.


Given a successful trajectory, we process it in reverse order from $t = T-1$ to $t = 0$. At each timestep, we maintain two critical pose configurations: the current robot pose $p_t$ and the subsequent pose $p_{t+1}$. These poses are used to compute two relative goal positions ${x_{t,\mathrm{\text{mirror}}}^g}$ and ${x_{t+1,\mathrm{\text{mirror}}}^g}$, which represent the goal in the robot's local coordinate frame at their respective timesteps. The transformed goal positions are then incorporated into new state representations:

\begin{equation}
x_{t,\mathrm{\text{mirror}}} = [L_t, {x_{t,\mathrm{\text{mirror}}}^g}, v_t, \omega_t]
\end{equation}
\begin{equation}
x_{t+1,\mathrm{\text{mirror}}} = [L_{t+1}, {x_{t+1,\mathrm{\text{mirror}}}^g}, v_{t+1}, \omega_{t+1}]
\end{equation}

The reward function for mirrored experiences follows:

\begin{equation}
\begin{aligned}
r_{t,\mathrm{\text{mirror}}} = 
\begin{cases}
r_{s}, & \text{if $\|{x_{t,\mathrm{\text{mirror}}}^g}\|_2 < \epsilon$,}\\
c_{1}\left(d_{t}^{g}-d_{t+1}^{g} \right), & \text{if $\|{x_{t,\mathrm{\text{mirror}}}^g}\|_2 \geq \epsilon$}.
\end{cases}
\end{aligned}
\end{equation}
where $r_s = 10.0$ is the success reward, $\epsilon = 0.2$ is the goal reaching threshold, $c_1 = 2.0$ is the progress scaling factor. 
Then the mirrored state, action and reward are stored in the standard training buffer $\mathcal{B}$ for the learning process.

\begin{figure}[t]
	\centering
	\includegraphics[width=0.42\linewidth]{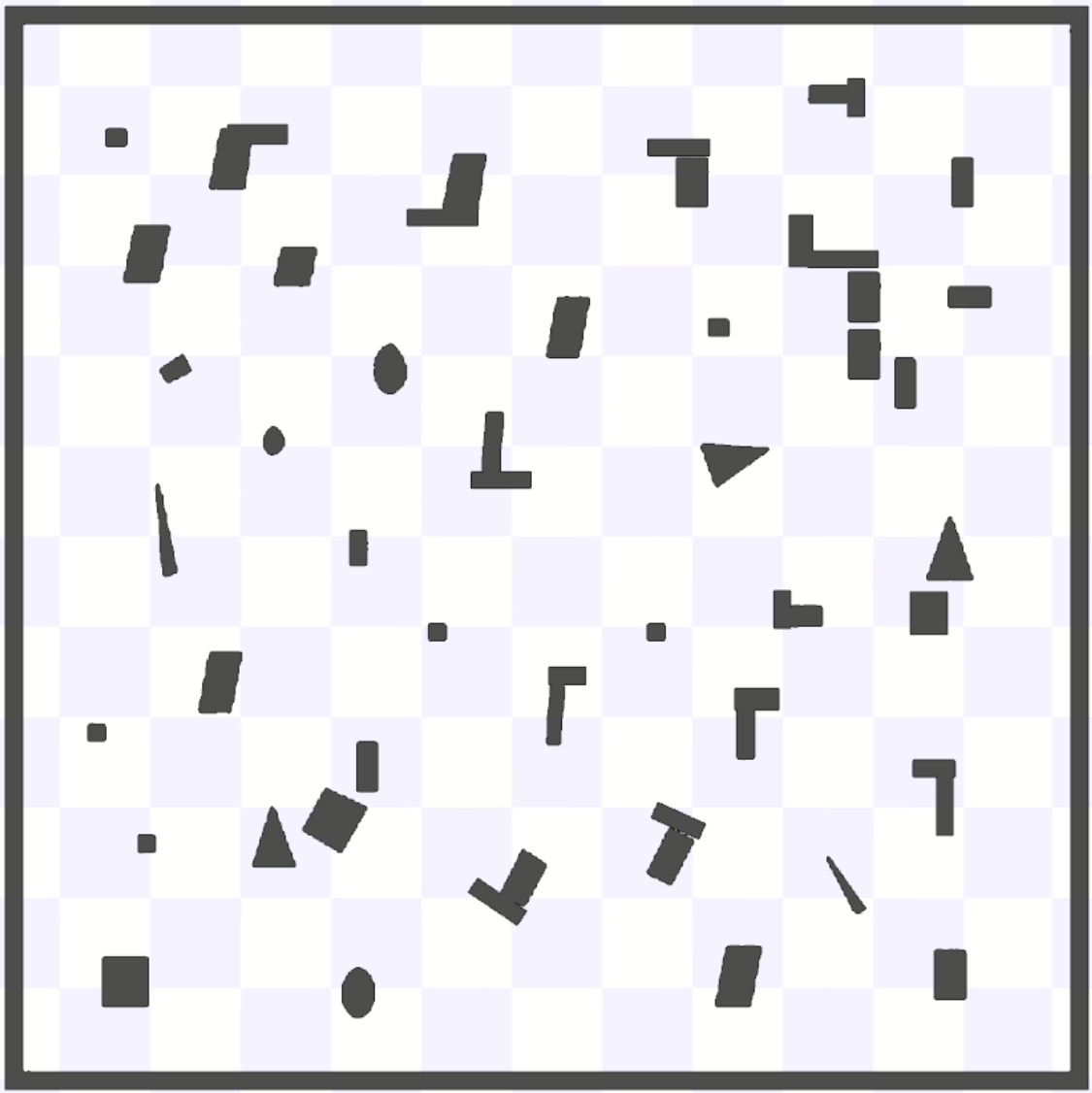}
	\caption{Training environment in \textit{ROS Stage} \cite{Stage_rosWiki}.}
	\label{training_env}
\end{figure}

\subsection{Proposed Network Structure}

Our proposed architecture employs the Soft Actor-Critic (SAC) framework for deep reinforcement learning navigation. The network (shown in Fig. \ref{network}), leveraging a comprehensive replay buffer composition with both standard and mirror-augmented experiences, processes multimodal sensory inputs to generate optimal navigation actions. The observation input comprises processed laser scans $q_1,...,q_m$, target position information ($d_t^g, \varphi_t^g$) (relative distance and angle), and robot velocity ($v_t, \omega_t$). Similar to \cite{zhang2022ipaprec}, $q_i$ is formulated as:

\begin{equation}
q_i = \frac{1}{l_i + \beta}
\end{equation}
where $l_i$ represents the minpooled LiDAR reading and $\beta$ is a trainable parameter. This transformation compresses the range of long-distance values while amplifying the range of short-distance values, thereby focusing the network's attention on immediate collision risks. 
The actor network processes the concatenated input information through four fully-connected layers with Leaky ReLU activation functions, culminating in a Gaussian Mixture Model output layer. This architecture enables the network to capture complex, multimodal action distributions necessary for sophisticated navigation behaviors.

To facilitate stable learning across environments of varying complexity, we implement an adaptive curriculum strategy. Training begins in a single environment, with new environments being unlocked when the success rate in current environments exceeds 70\%. Environment selection probability is computed as $p(\mathrm{env}_i) = (1 - \mu_i)/\sum_{j} (1 - \mu_j)$, where $\mu_i$ represents the mean success rate. This ensures focused training on challenging scenarios while maintaining performance in mastered environments. The curriculum expands in a structured pattern, unlocking adjacent environments both horizontally and vertically in our grid-based environment setup, effectively balancing exploration of new challenges with consolidation of learned skills. The algorithm for training MAER-Nav is depicted
in Algorithm \ref{algorithm-MAER-Nav}.

\begin{algorithm}[t]
\SetAlgoLined
 Initialize parameters of policy network $\theta$, critic networks $\phi_1$ and $\phi_2$, total training steps $T_\mathrm{\text{total}}=0$ and empty replay buffers $\mathcal{B}$ and $\mathcal{B}_\mathrm{\text{mirror}}$\;
 \For{\upshape{episode}$=1$, 2, \ldots}
 {
    Select environment with probability $p \propto (1-\bar{\rho})$ where $\bar{\rho}$ are mean success rates\;
    
    \While{$t<T_{\mathrm{\text{max}}}$ \upshape{\textbf{and} not terminate}}
    {
      \eIf{\upshape{training}}{Sample action $a_t\sim\pi_{\theta}(x_t)$\;}
      {Sample a random action $a_t$\;}
      Execute $a_t$ in simulation\;
      Obtain next observation $x_{t+1}$ and robot pose $p_{t+1}$, reward $r_t$ , and the termination signal $d_t$\;
      Store $\{x_t,a_t,r_t,x_{t+1},d_t\}$ in $\mathcal{B}$, store $\{(x_t, a_t, r_t, x_{t+1}, d_t, p_t, p_{t+1})\}$ in $\mathcal{B}_\mathrm{\text{mirror}}$\;
      $x_t\leftarrow x_{t+1}$, $p_t\leftarrow p_{t+1}$, $t\leftarrow t+1$\;
    }
    
    \If{goal\_reached $= 1$}
    {
      \For{each transition in episode}
      {
        Create negative action $a' = -a$, mirrored goal transition and robot pose from $\mathcal{B}_\mathrm{\text{mirror}}$, store them in $\mathcal{B}$\;
        Clear $\mathcal{B}_\mathrm{\text{mirror}}$\;
      }
    }
    
    \If{\upshape{Training}}
    {
      \For{$j=1$ to $t$}
      {
        $T_\mathrm{\text{total}}\leftarrow T_\mathrm{\text{total}}+1$\;
        Sample a minibatch from replay buffer $\mathcal{B}$\;
        Update critic networks $\phi_1, \phi_2$\;
        Update policy network $\theta$\;
      }
      
    }
 }
\caption{Training of MAER-Nav}
\label{algorithm-MAER-Nav}
\end{algorithm}

\section{Robot Training in Simulation}

The training of MAER-Nav was conducted in \textit{ROS Stage} \cite{Stage_rosWiki} simulation environment. The simulation setup consists of a differential-drive robot equipped with a 2D LiDAR sensor that maintains specifications consistent with real-world hardware, featuring a 360$^{\circ}$ field of view (FOV), 0.216$^{\circ}$ angular resolution (1667 laser beams), and a maximum sensing range of 30 meters. The robot operates with maximum linear and angular velocities of 0.5 m/s and $\pi/2$ rad/s, respectively, with a control frequency of 10 Hz. Each training episode is limited to 400 time steps to ensure consistent learning conditions.
Starting with an initial map size of 20$\times$20 m$^2$ (shown in Fig. \ref{training_env}), the training environments progressively decrease in size across a 5$\times$5 grid of maps, creating 25 environments with varying difficulty levels. 

To evaluate the effectiveness of our proposed Mirror-Augmented Experience Replay mechanism, we implemented a baseline version (MAER-Raw) that excludes the mirror-augmented replay buffer while maintaining all other architectural components. This baseline implementation serves as a control to isolate the impact of bidirectional motion learning on navigation performance.

\section{Simulation Test Results and Analysis}

\begin{figure}[t]
	\centering
    \subfloat[Corridor]{
\centering\includegraphics[width=0.23\linewidth]{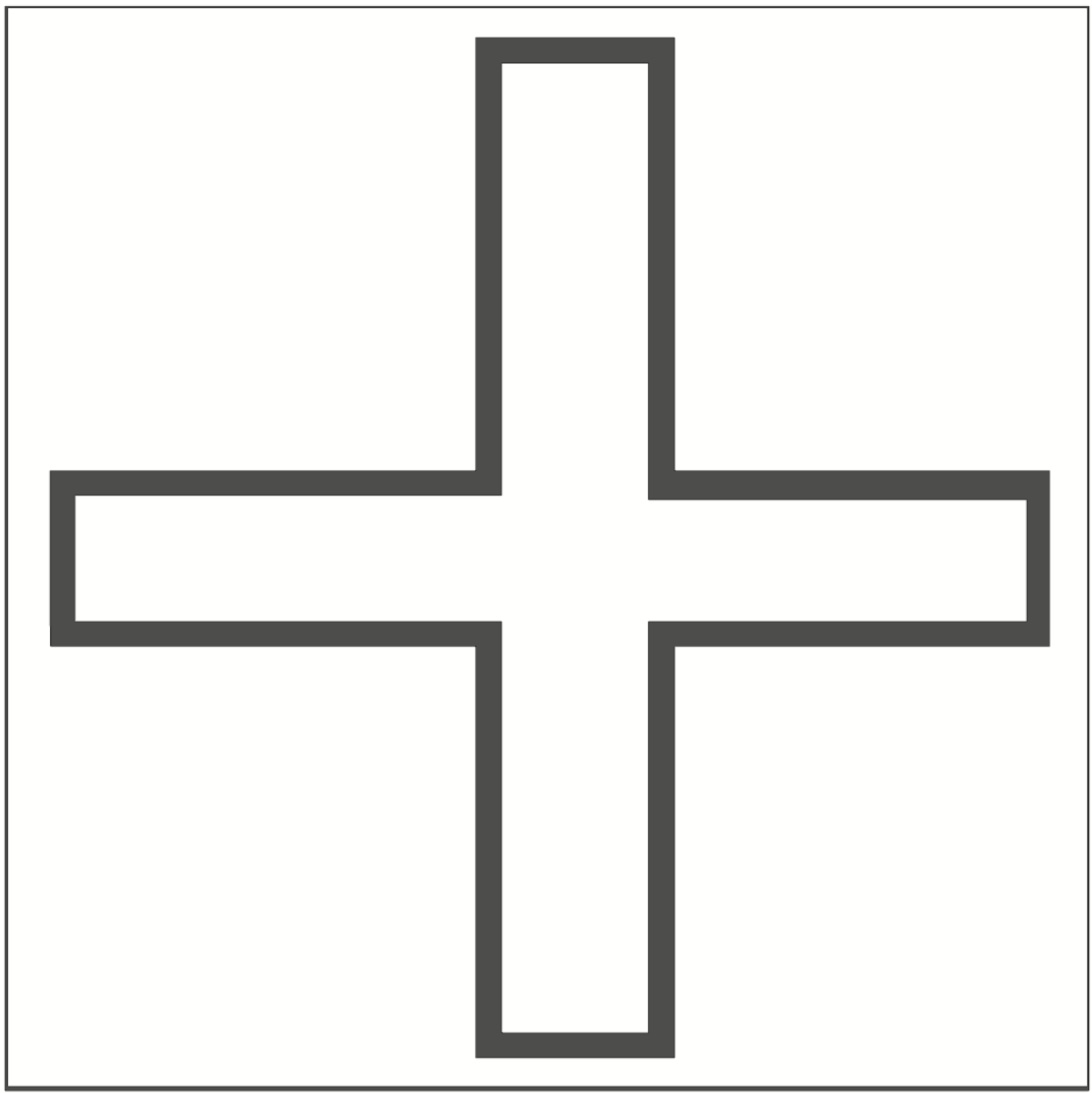}\label{s1}}
    \subfloat[Wall]{
\centering\includegraphics[width=0.23\linewidth]{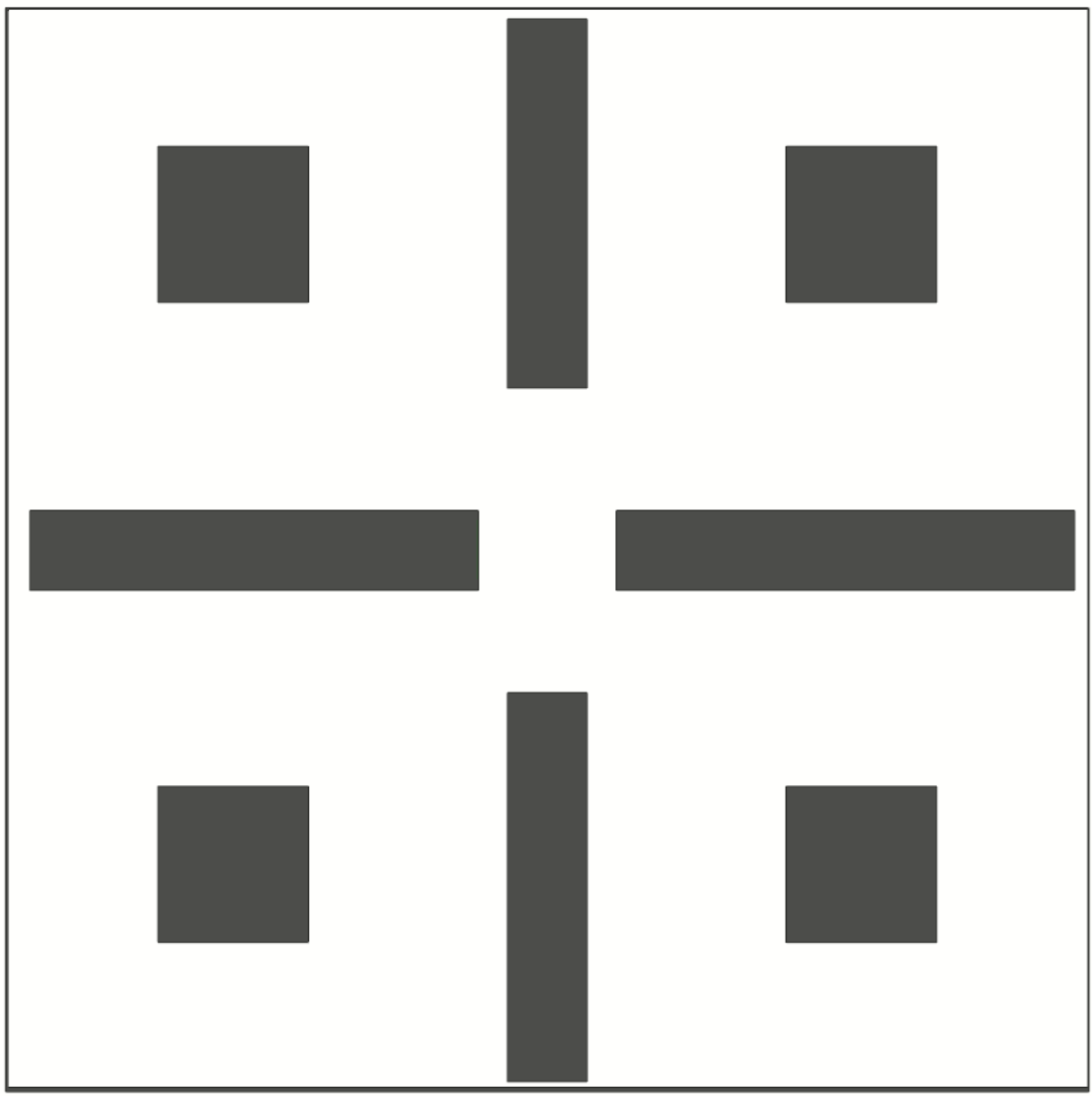}\label{s2}}
    \subfloat[SEnv1]{
\centering\includegraphics[width=0.23\linewidth]{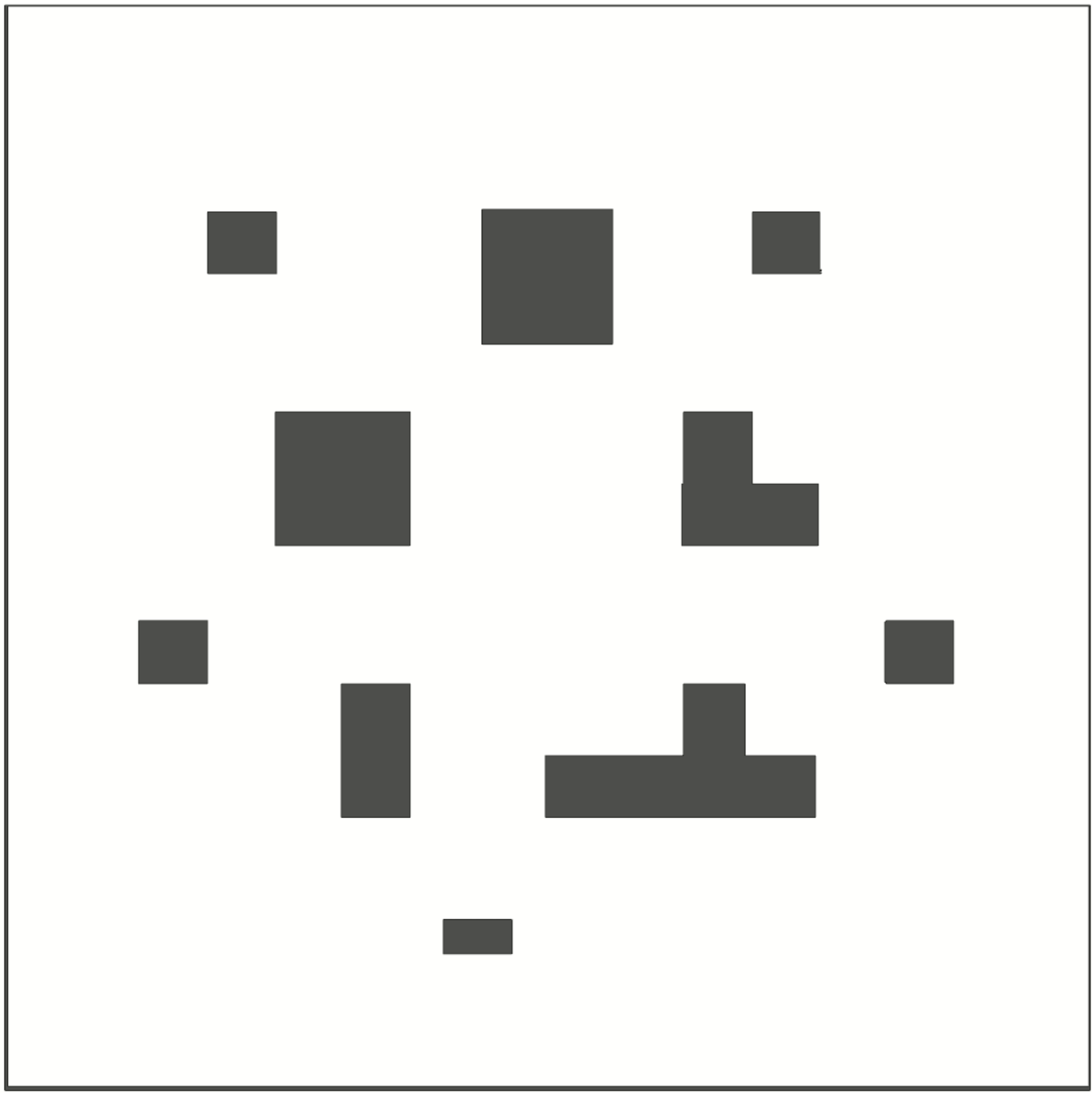}\label{s3}}
    \subfloat[SEnv2]{
\centering\includegraphics[width=0.23\linewidth]{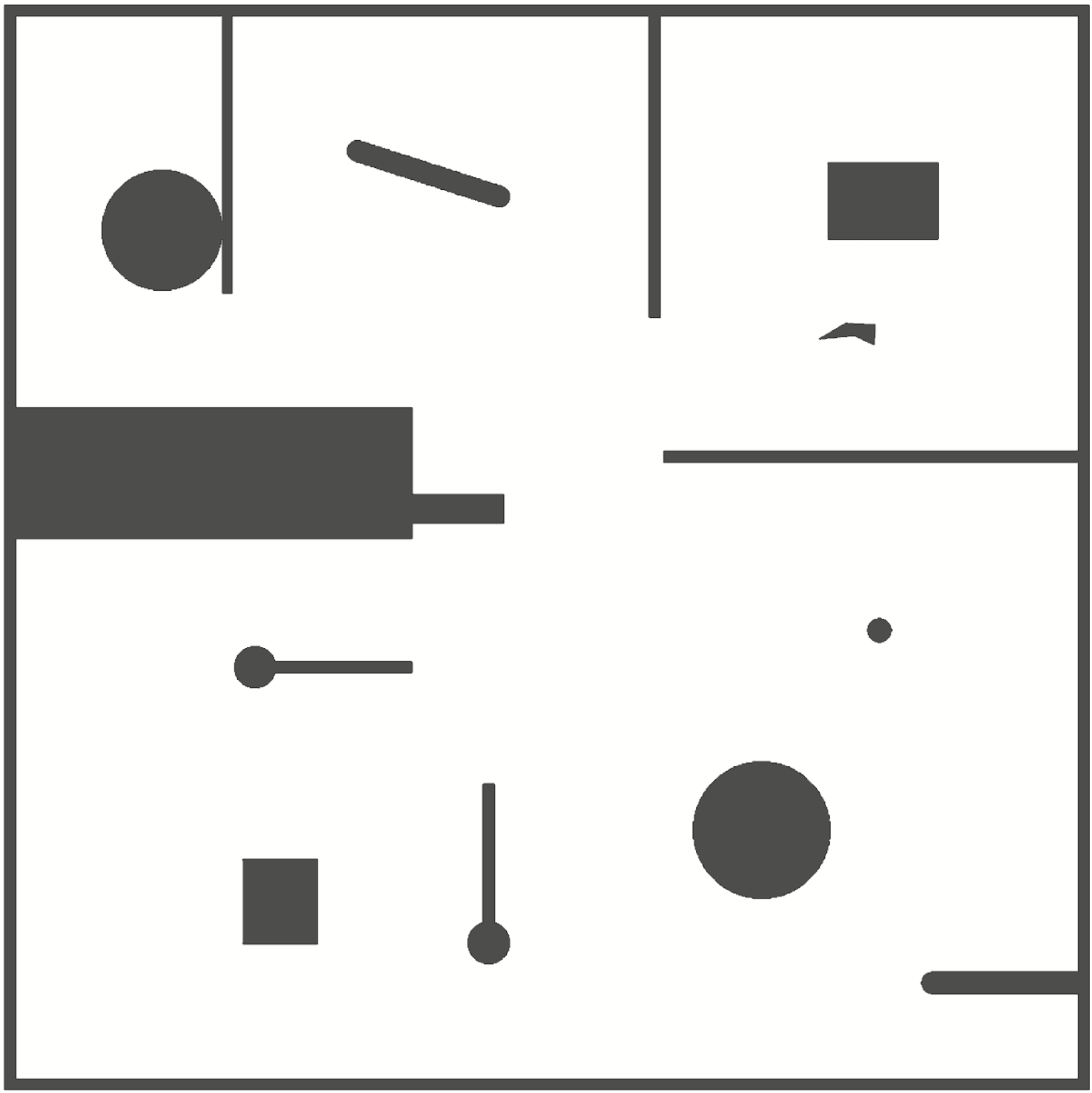}\label{s4}}
	
	\caption{Four unseen
test environments: Narrow Corner Corridor, Multiple Walls, SEnv1 and SEnv2 for simulation test.}
	\label{sim_test}
\end{figure}

To evaluate the performance of our trained algorithm, we conducted extensive simulation-based testing in \textit{ROS Stage}. The robot was represented by a black rectangular box with a red marker at the front indicating the LiDAR position and forward direction. The robot's dimensions were set to match our real-world platform, with a length of 0.62 m and width of 0.64 m, operating at a maximum linear velocity of 0.5 m/s.

\begin{figure}[t]
	\centering
	\includegraphics[width=\linewidth]{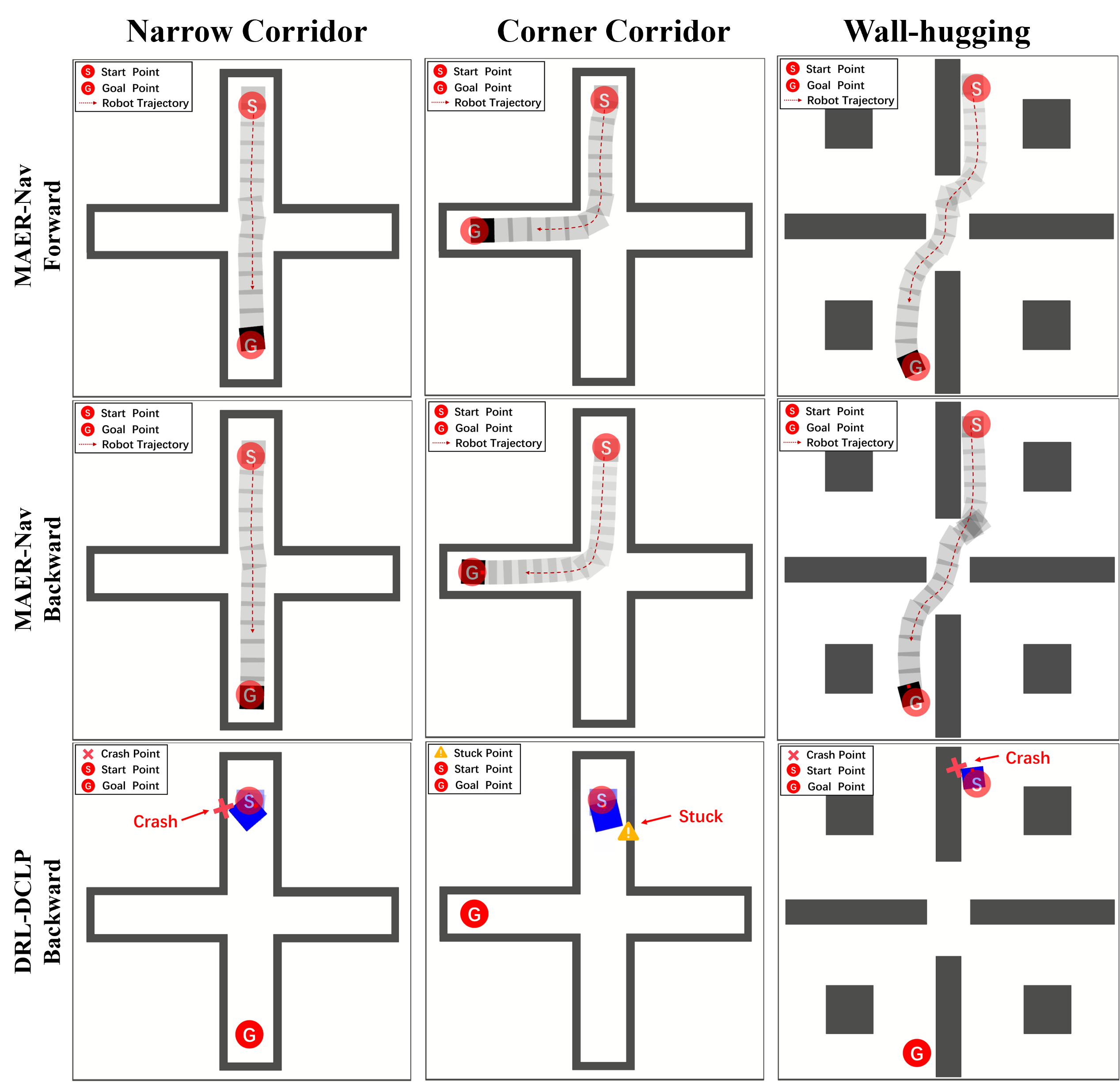}
	\caption{Trajectory comparison in corridor and wall scenarios: successful forward and backward navigation trajectories of MAER-Nav (black robot) versus the failed backward trajectories of DRL-DCLP (blue robot). `S' and `G' represent the robot's start point and goal point, respectively. Start, goal, crash and stuck positions, along with the robot's trajectories, are indicated in the legend.}
	\label{shiziwallall}
\end{figure}

Our evaluation utilized two categories of test environments. The first category consisted of classical challenging scenarios designed to compare MAER-Nav against MAER-Raw and DRL-DCLP \cite{DRL-DCLP}, one of the state-of-the-art approaches in recent robotic navigation research. The first two maps included a narrow cross-shaped corridor and a multiple walls environment, as shown in Fig. \ref{s1} and \ref{s2}, respectively. Both maps measured 8$\times$8 m$^2$. In the corridor scenario, the robot's starting position was set at the uppermost point of the corridor, with two separate goal positions at the leftmost and bottommost ends. For the wall environment, we positioned the robot in close proximity to the wall to test its capability in extreme situations. Among the above scenarios, any initial rotational movement would result in an immediate collision, making it particularly challenging for navigation. 
We evaluated the bidirectional navigation capabilities of MAER-Nav and compared it with the backward motion capabilities of DRL-DCLP. As shown in Fig. \ref{shiziwallall}, the black vehicle represents MAER-Nav, while the blue vehicle represents DRL-DCLP. Our method demonstrated excellent navigation performance and trajectory symmetry across three challenging scenarios: Narrow Corridor, Corner Corridor, and Wall-hugging. Notably, when the robot's initial pose was set backward, DRL-DCLP invariably experienced crashes or became stuck in these scenarios due to the excessively narrow initial environment that prevented orientation adjustment. This represents a classic challenge for traditional DRL-based methods. In contrast, our method effectively handled these situations with smooth and stable backward trajectories.
The second category comprised more complex environments, as illustrated in Fig. \ref{s3} and Fig. \ref{s4}, with map dimensions of 12$\times$12 m$^2$. Each map contained 50 point-to-point navigation tasks, where the robot was required to navigate from a designated start position to a goal location within a maximum number of allowed steps. The objective was to complete these tasks in minimal time while maintaining high navigation performance.

The evaluation of the algorithm includes several performance metrics: average success rate (SR), average collision rate (CR), average timeout rate (TR), average episode steps (AES*), and maximum average navigation score (MANS, as MEAN ± SD). The AES* metric specifically considers only the episodes where all three methods succeeded, computing the average length across this intersection of successful episodes. The MANS metric is defined as:
\begin{equation}
S = \begin{cases}
1 - \frac{2T_s}{T_\mathrm{\text{max}}}, & \text{if success}, 
\\
\
-1, & \text{otherwise},
\end{cases}
\end{equation}
where $T_s$ denotes the number of steps required by the agent to complete navigation. This metric offers a holistic assessment that incorporates both the time spent navigating and successful task fulfillment, serving as an accurate indicator of the agent's navigational efficiency.

Table I presents a comprehensive quantitative assessment of navigation performance across three methods evaluated in two simulation environments. In SEnv1, MAER-Nav demonstrated superior performance with a 100\% success rate compared to DRL-DCLP (80\%) and MAER-Raw (70\%). MAER-Nav recorded no collision events (0\%) in contrast to the 16\% and 12\% rates observed with DRL-DCLP and MAER-Raw respectively. Furthermore, MAER-Nav exhibited zero timeout incidents and achieved more efficient navigation with fewer steps (201.31) than both DRL-DCLP (207.53) and MAER-Raw (215.81).
In SEnv2, MAER-Nav maintained its performance advantage with a 98\% success rate, substantially exceeding MAER-Raw (86\%) and DRL-DCLP (80\%). The maximum average navigation score for MAER-Nav (0.29 ± 0.28) was significantly higher than those of DRL-DCLP (0.08 ± 0.56) and MAER-Raw (0.14 ± 0.51), with notably lower variance. These results quantitatively validate MAER-Nav's enhanced path efficiency and operational stability across complex navigation scenarios.

\begin{table}[t]
\centering
\caption{\label{tab:simulation results}simulation performance of different methods in SENV1 and SENV2.}
\renewcommand\arraystretch{1.5}
\setlength{\tabcolsep}{1.4mm}
\begin{tabular}{cccccc}
\hline
\hline
Method   & SR(↑)         & CR(↓)         & TR(↓)      & AES*(↓)         & MANS(↑)                  \\ \hline
         & \multicolumn{5}{c}{SEnv1}                                                               \\ \cline{2-6} 
MAER-Raw & 70\%           & 12\%          & 18\%       & 215.81          & -0.17 ± 0.38         \\
DRL-DCLP & 80\%           & 16\%          & 4\%       & 207.53          & -0.11 ± 0.32         \\ \rowcolor{lightgray}
\textbf{MAER-Nav (Ours)} & \textbf{100\%} & \textbf{0\%} & \textbf{0\%} & \textbf{201.31} & \textbf{0.02 ± 0.20} \\ \hline
         & \multicolumn{5}{c}{SEnv2}                                                               \\ \cline{2-6} 
MAER-Raw & 86\%          & 14\%          & \textbf{0\%} & 128.54          & 0.14 ± 0.51          \\
DRL-DCLP & 80\%           & 20\%           & \textbf{0\%} & 131.32          & 0.08 ± 0.56          \\ \rowcolor{lightgray}
\textbf{MAER-Nav (Ours)} & \textbf{98\%} & \textbf{2\%} & \textbf{0\%} & \textbf{127.38} & \textbf{0.29 ± 0.28} \\
\hline
\hline
\end{tabular}
\end{table}

\section{Performance Validation in REAL-WORLD}

\begin{figure*}[t!]
	\centering\subfloat[REnv1]{
	\centering\includegraphics[width=0.151\linewidth]{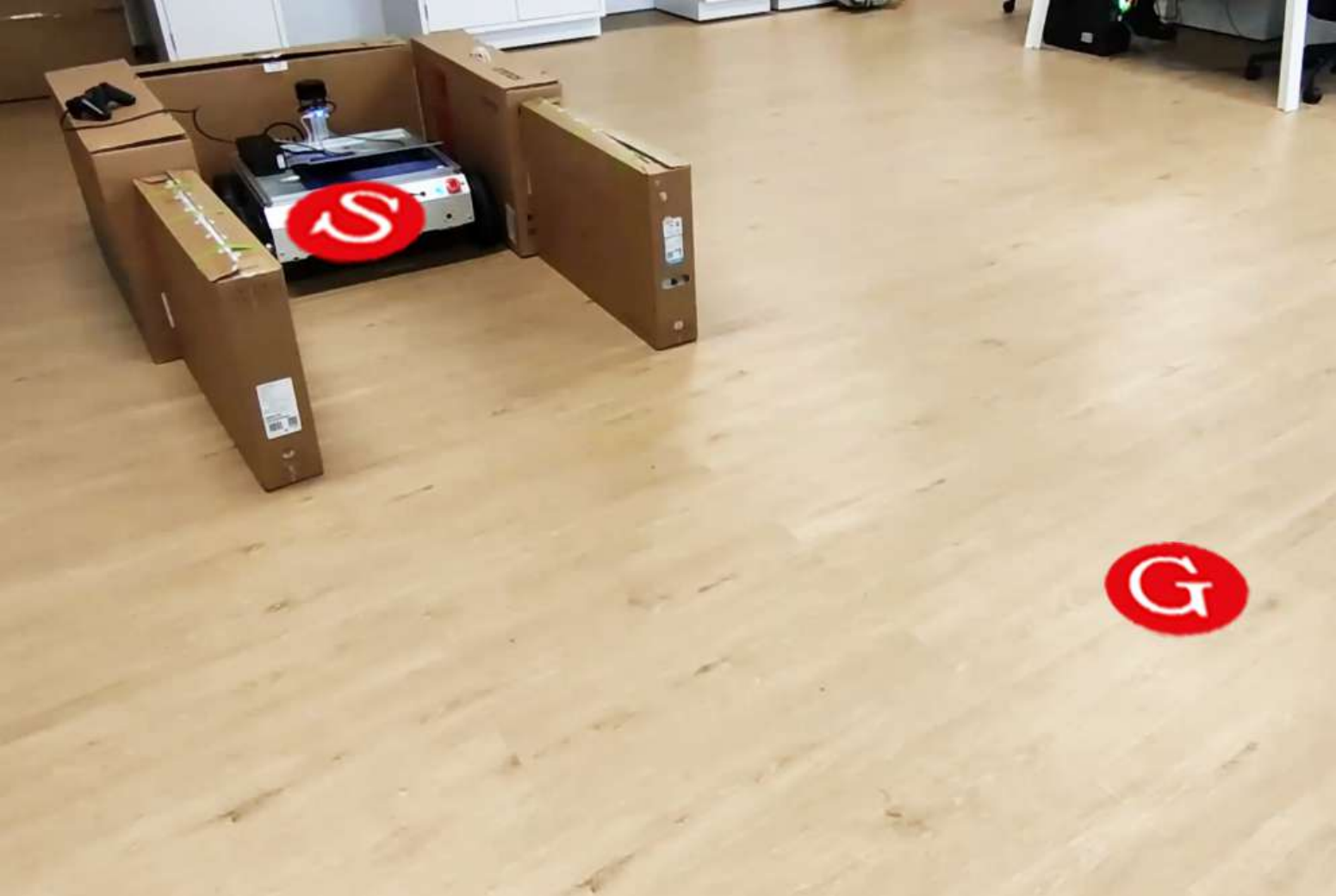}\label{r1}}
	 \hfill\subfloat[REnv2-1]{
	\centering\includegraphics[width=0.161\linewidth]{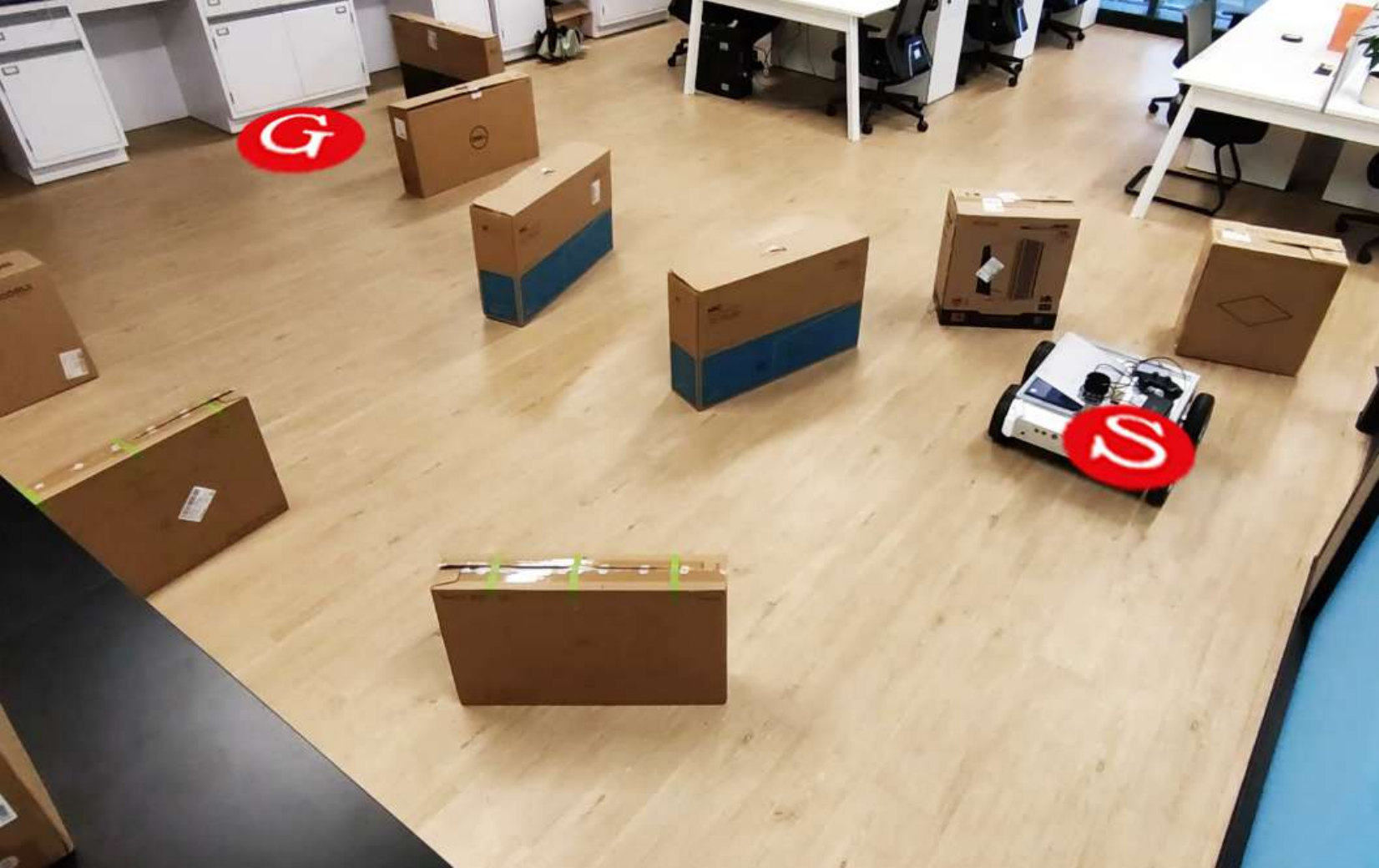}\label{r2-1}}
	 \hfill\subfloat[REnv2-2]{
	\centering\includegraphics[width=0.1665\linewidth]{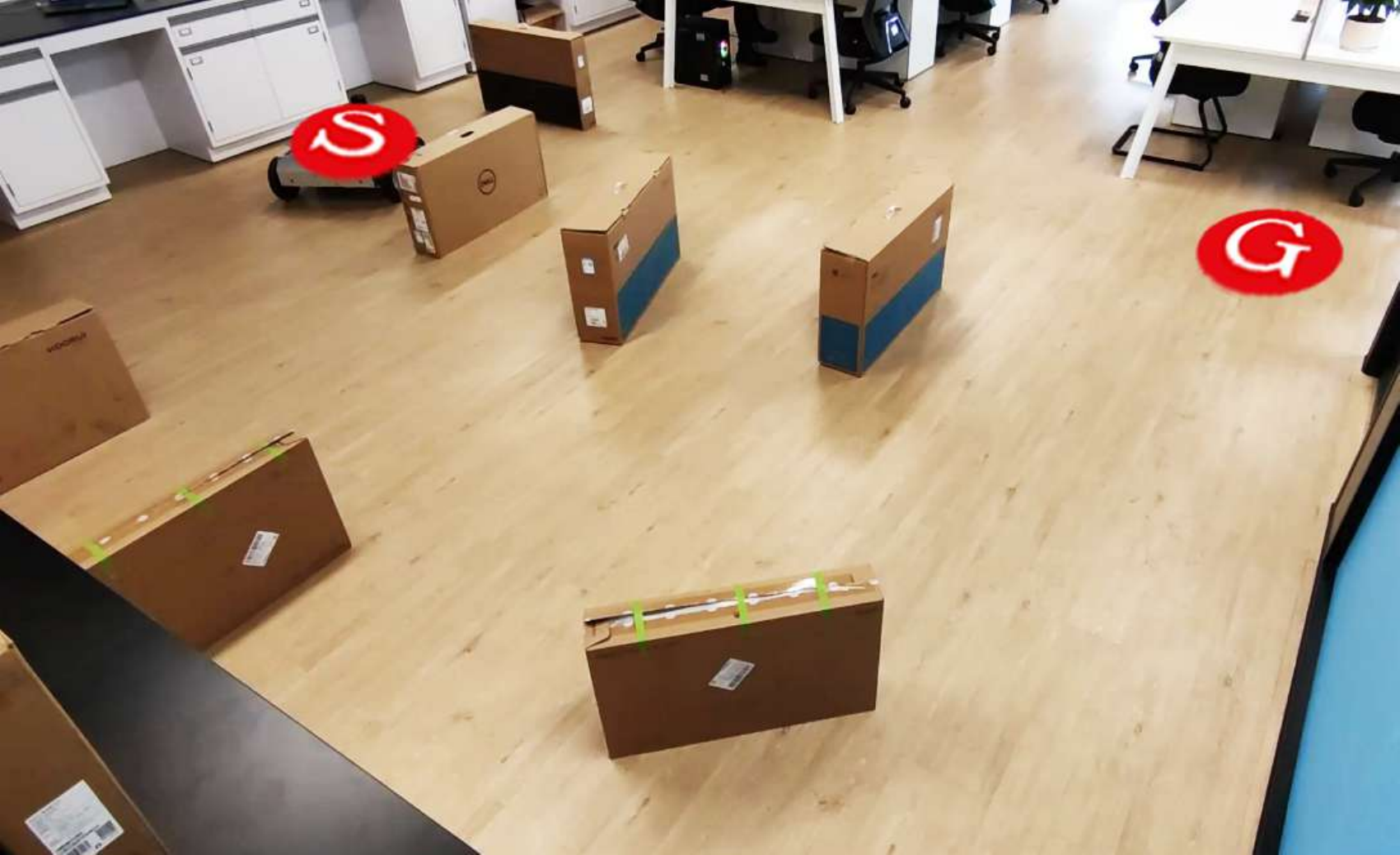}\label{r2-2}}
	 \hfill\subfloat[REnv3]{
	\centering\includegraphics[width=0.143\linewidth]{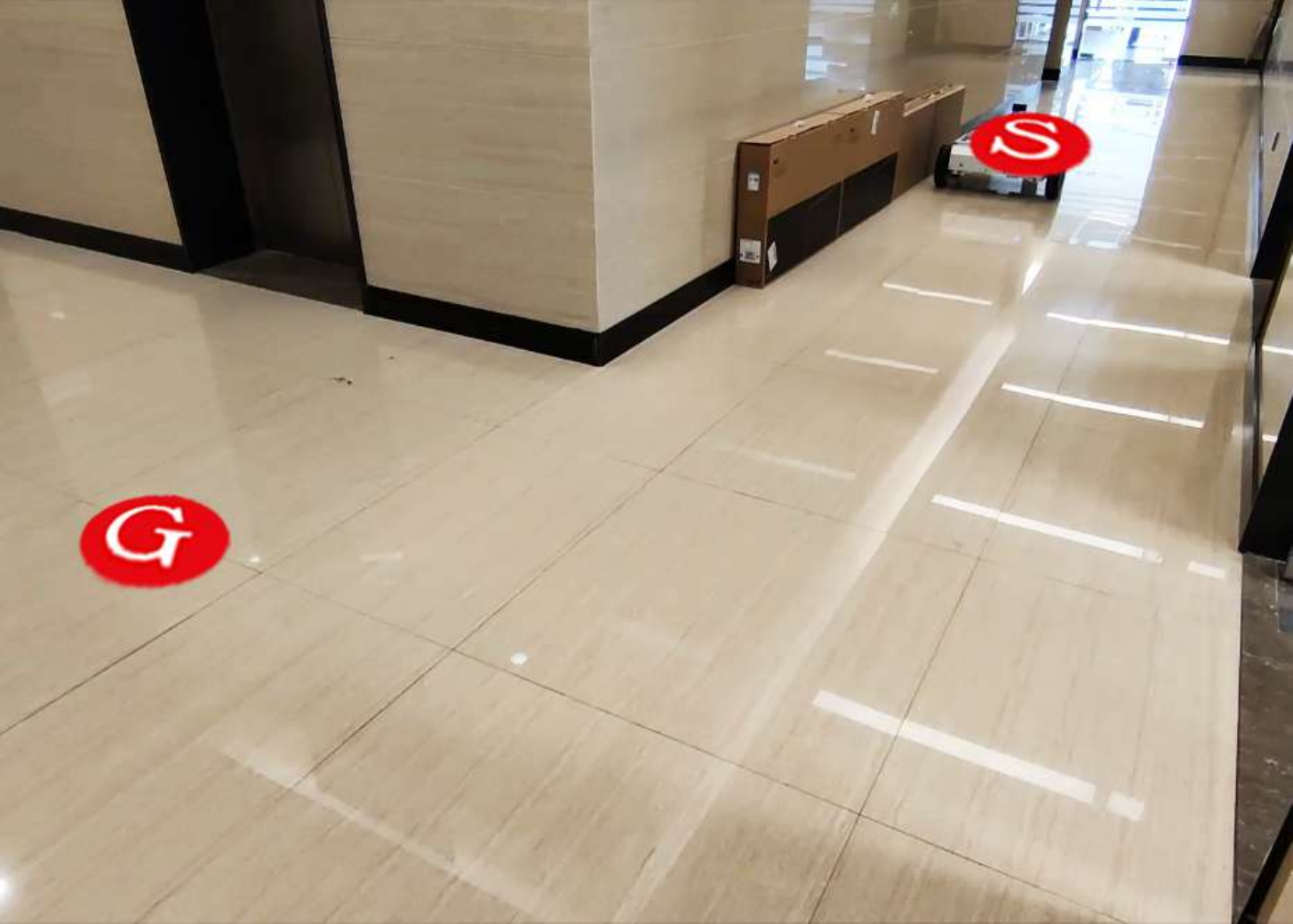}\label{r3}}
    \hfill\subfloat[REnv4]{
	\centering\includegraphics[width=0.181\linewidth]{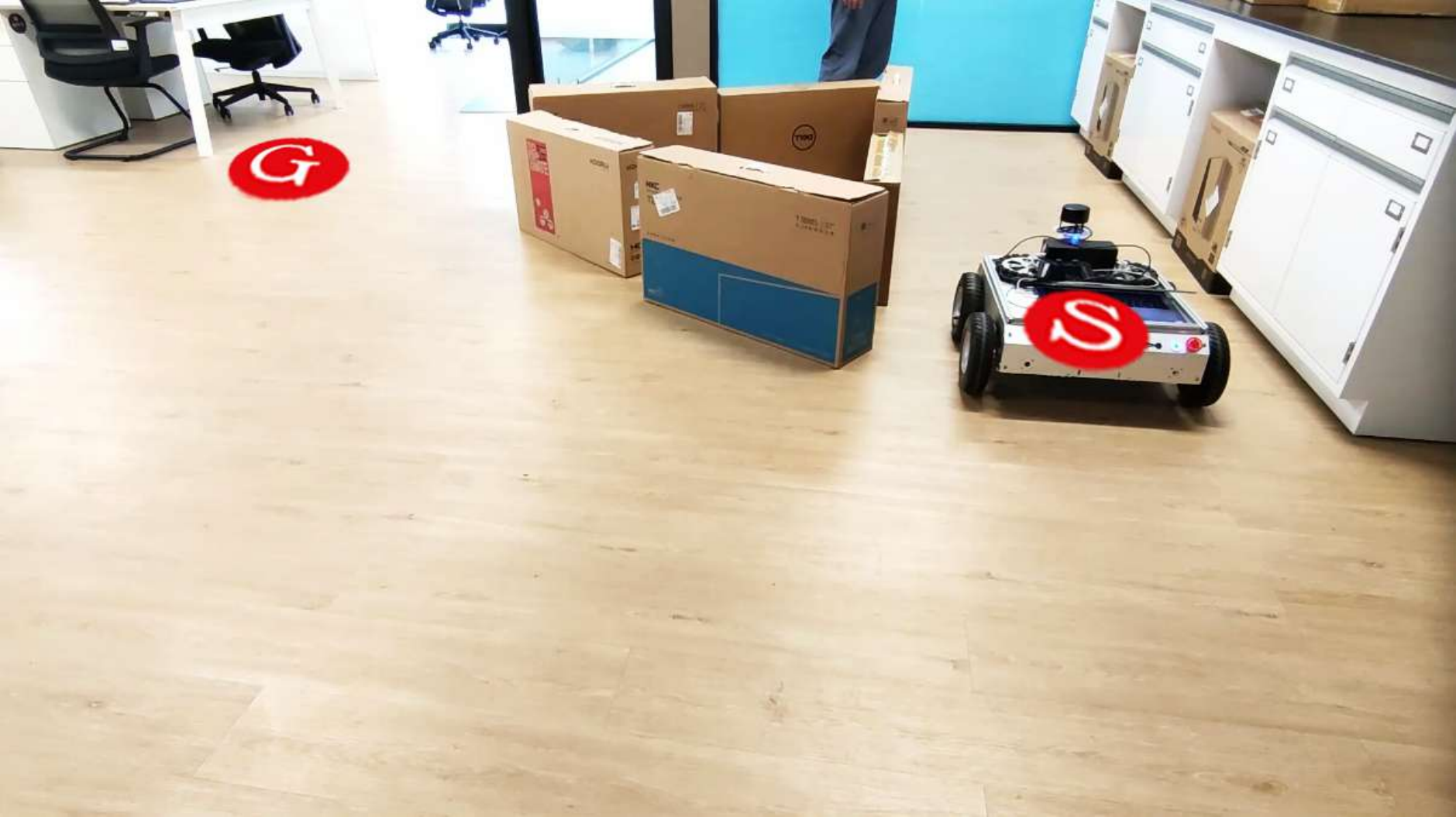}\label{r4}}
    \hfill\subfloat[Test robot]{
	\centering\includegraphics[width=0.155\linewidth]{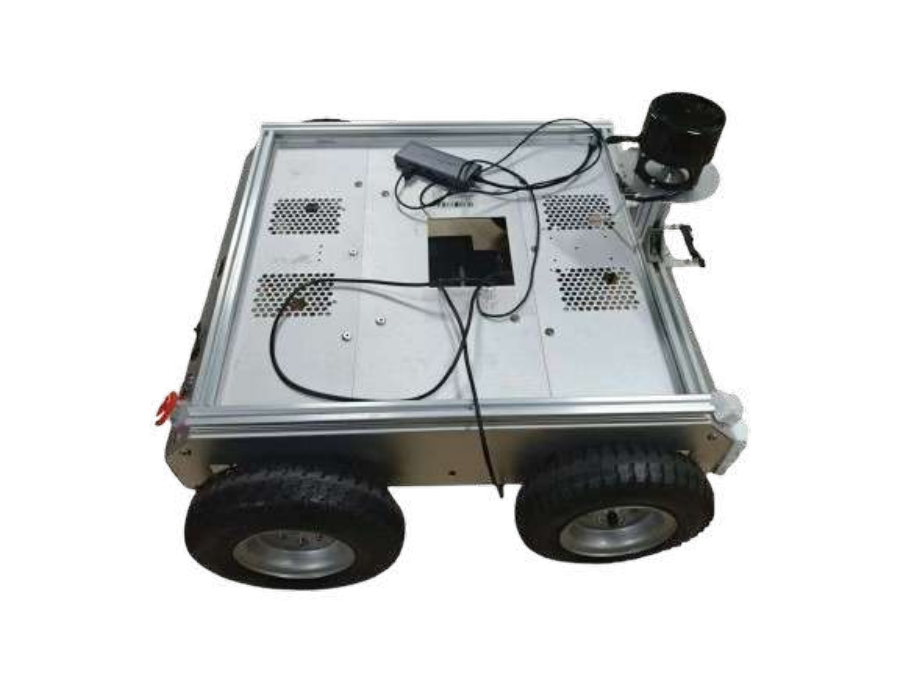}\label{robot}}
	\caption{Real-world test scenarios featuring the experimental robot, with start (`S') and goal (`G') positions marked.}
	\label{realtestenv}
\end{figure*}

\begin{figure*}[t]
	\centering\subfloat[MAER-Nav in REnv1]{
	\centering\includegraphics[width=0.245\linewidth]{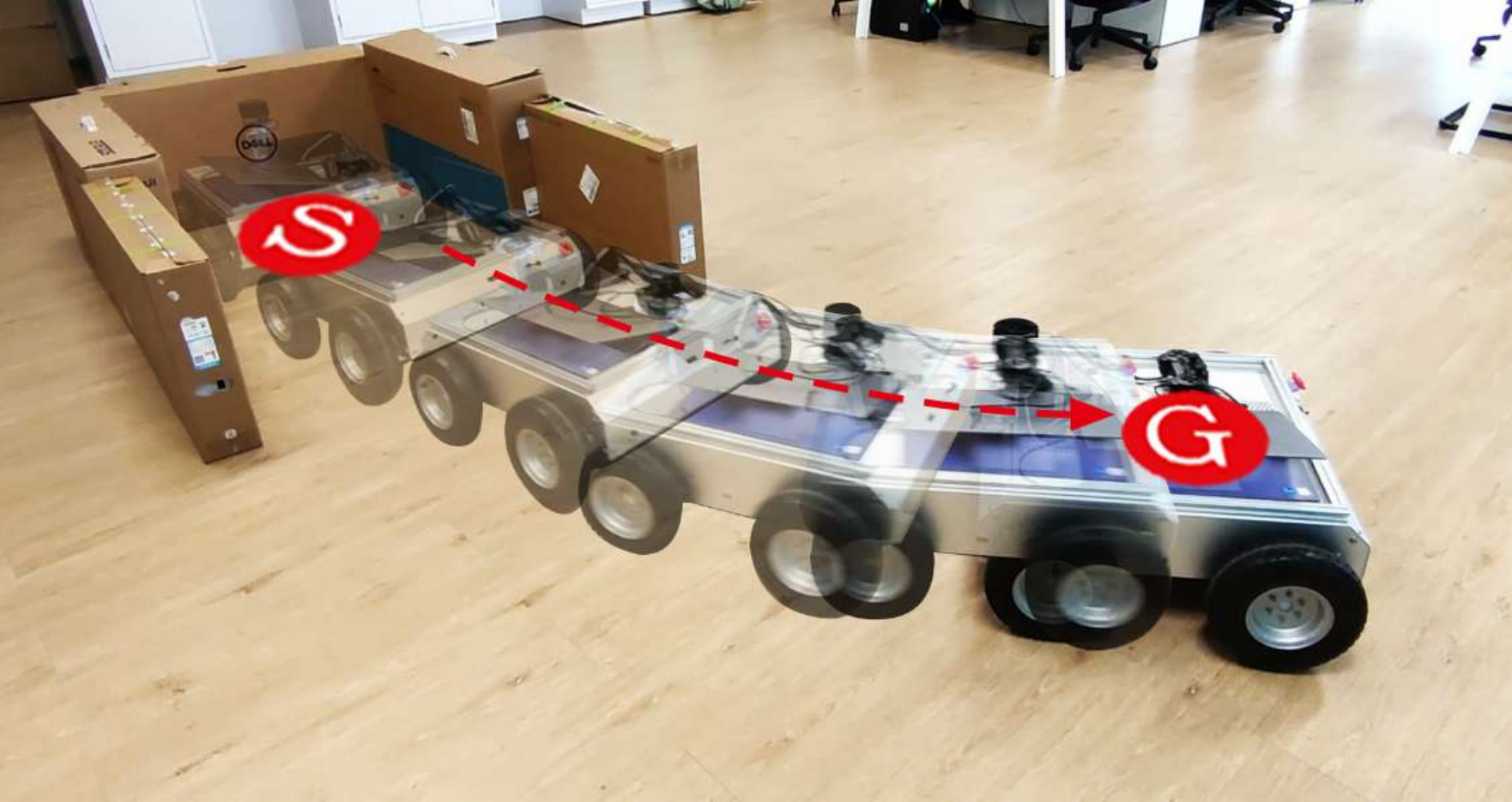}\label{maer-r1}}
	\hfill\subfloat[MAER-Nav in REnv2-1]{
	\centering\includegraphics[width=0.223\linewidth]{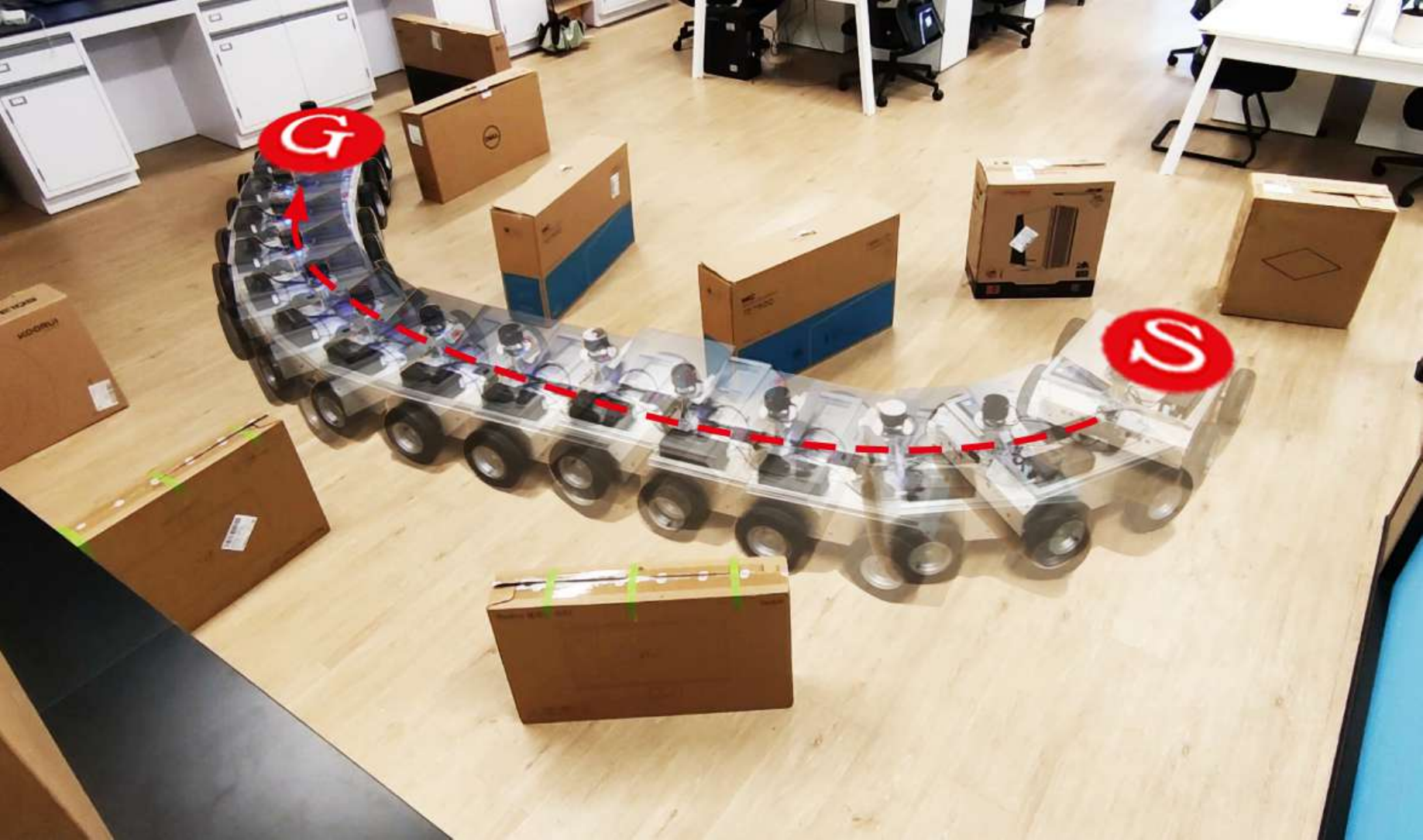}\label{maer-r2-1}}
    	\hfill\subfloat[MAER-Nav in REnv2-2]{
	\centering\includegraphics[width=0.218\linewidth]{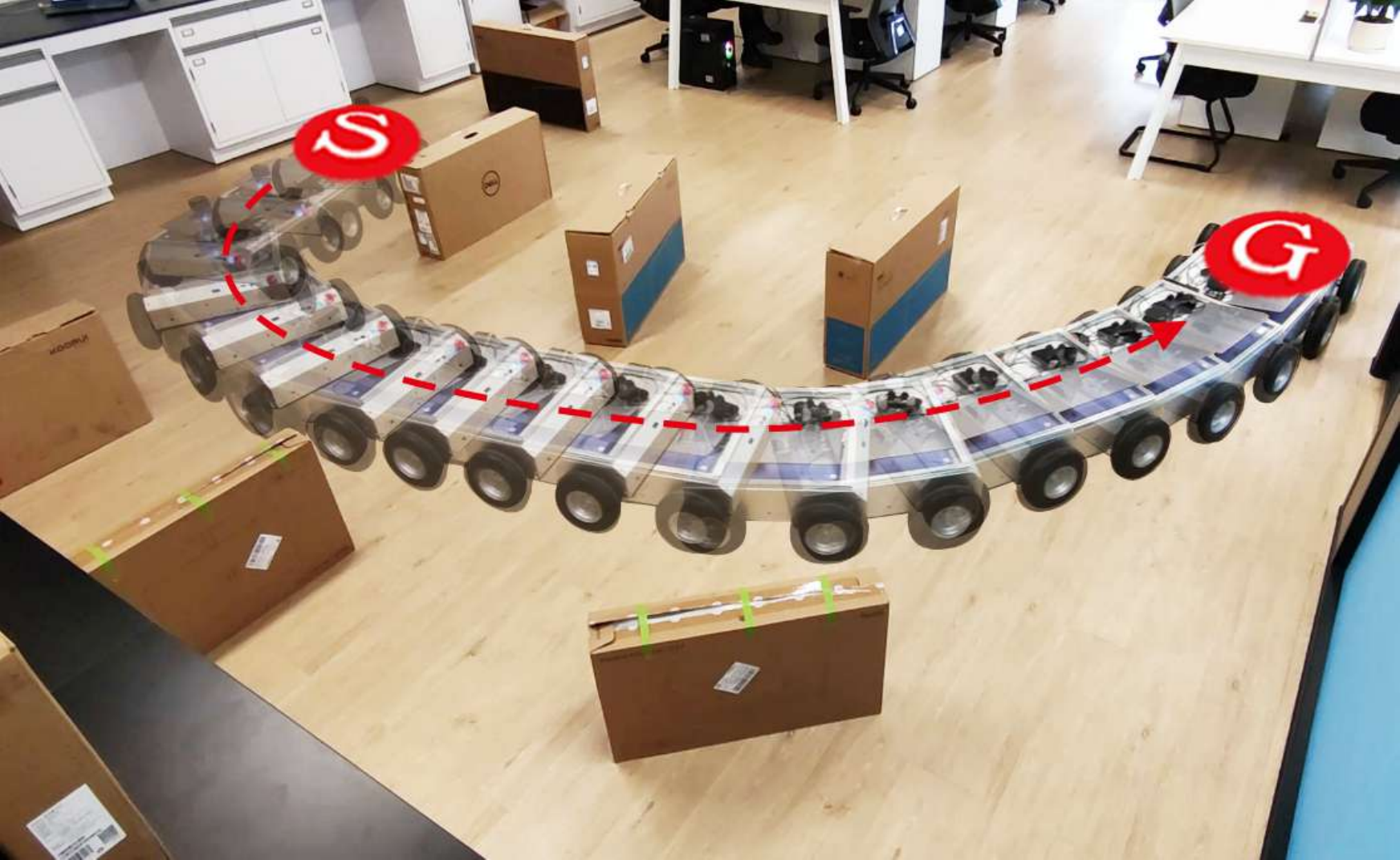}\label{maer-r2-2}}
    	\hfill\subfloat[MAER-Nav in REnv3]{
	\centering\includegraphics[width=0.242\linewidth]{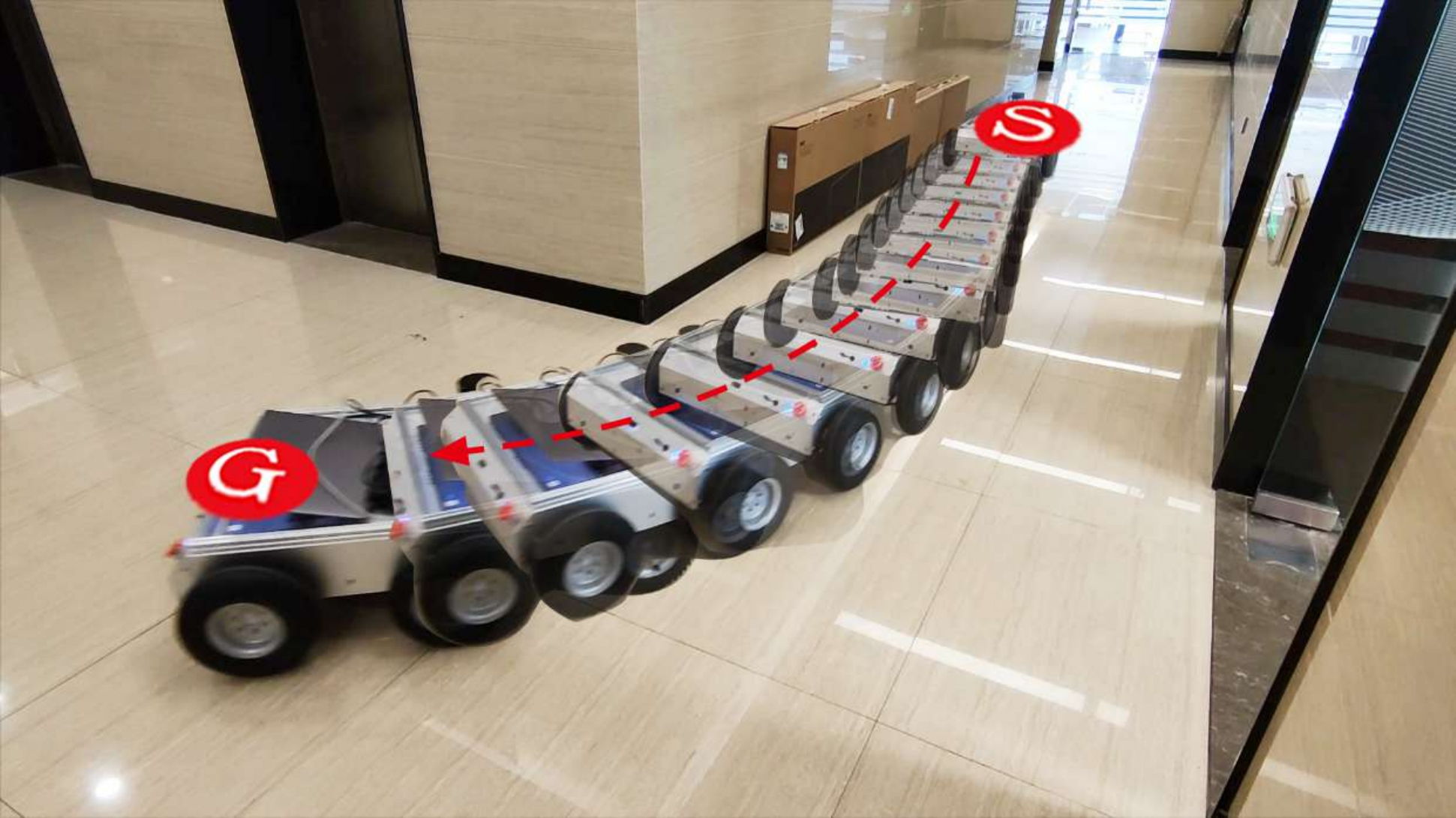}\label{maer-r3}}
	\\[-2ex]
	\subfloat[DRL-DCLP in REnv1]{
	\centering\includegraphics[width=0.247\linewidth]{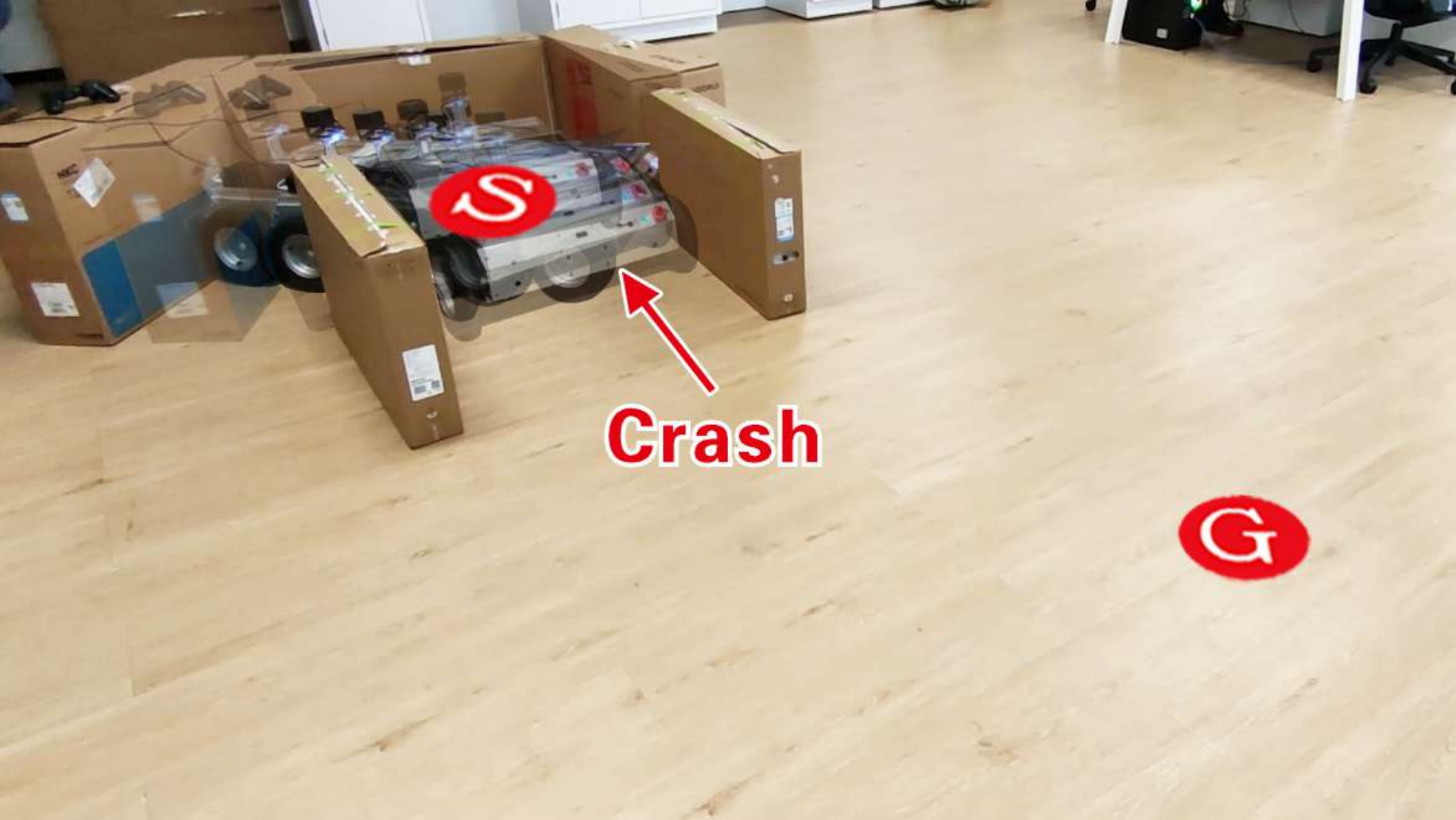}\label{dclp-r1}}
	\hfill\subfloat[DRL-DCLP in REnv2-1]{
	\centering\includegraphics[width=0.22\linewidth]{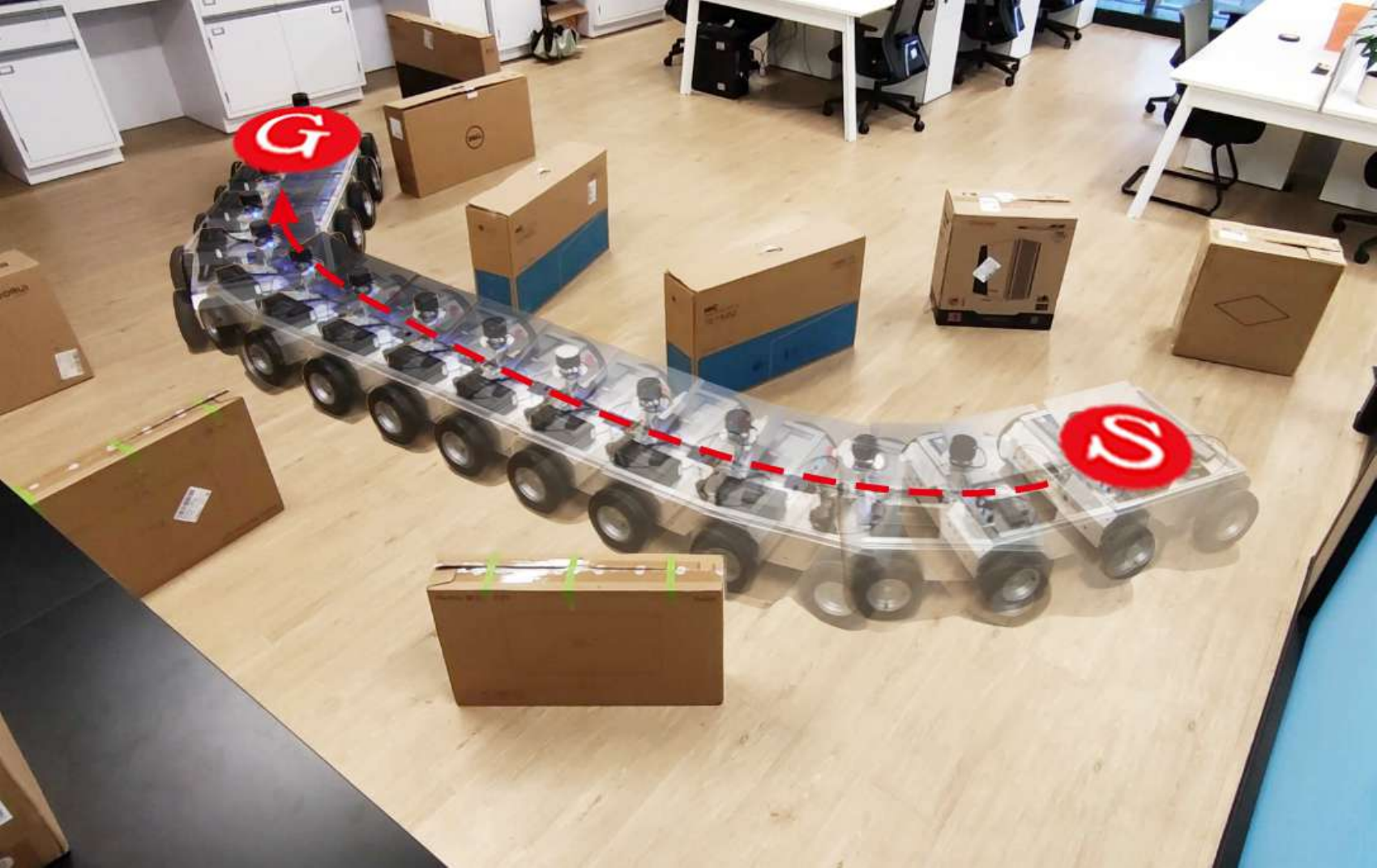}\label{dclp-r2-1}}
    	\hfill\subfloat[DRL-DCLP in REnv2-2]{
	\centering\includegraphics[width=0.22\linewidth]{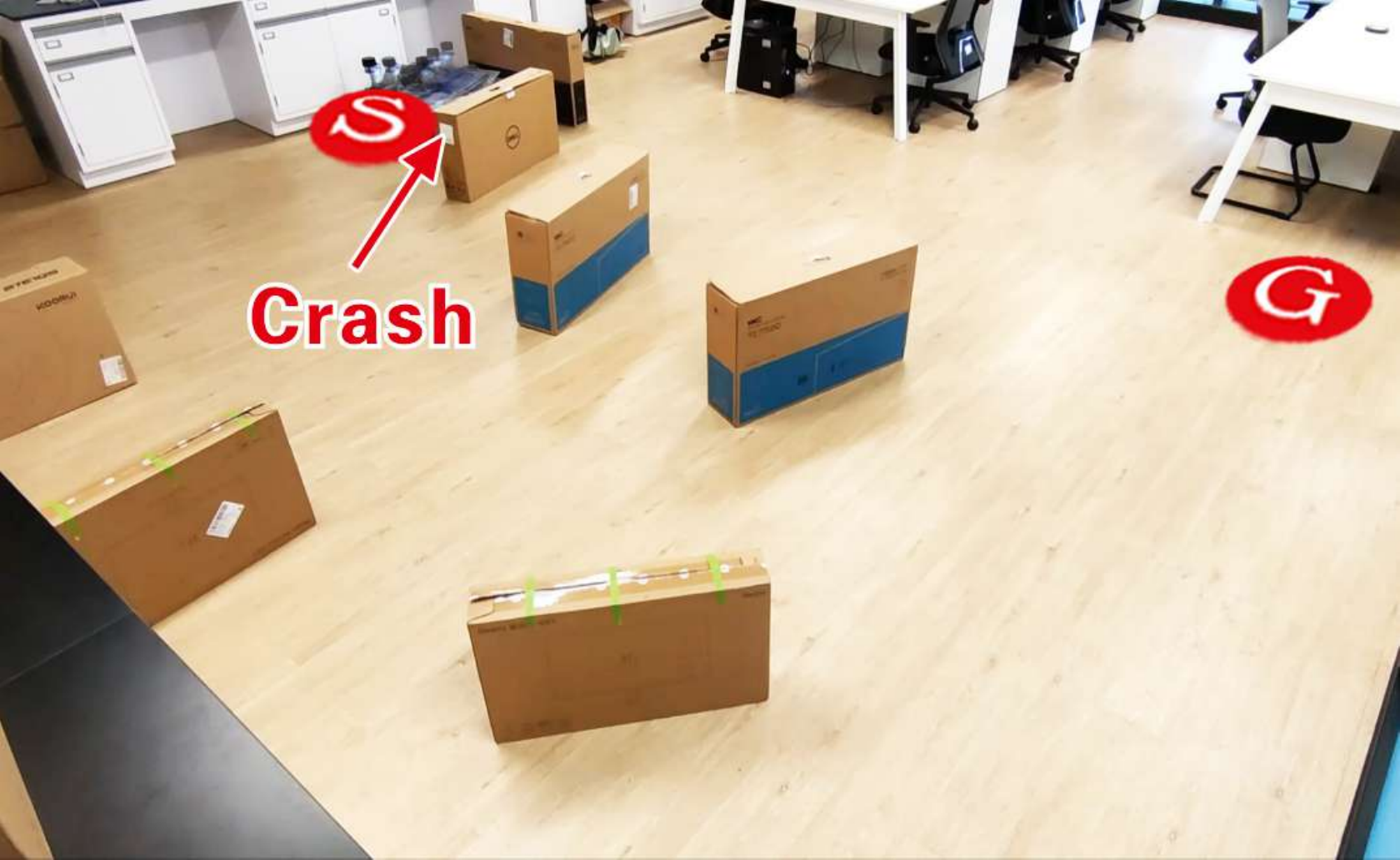}\label{dclp-r2-2}}
    	\hfill\subfloat[DRL-DCLP in REnv3]{
	\centering\includegraphics[width=0.245\linewidth]{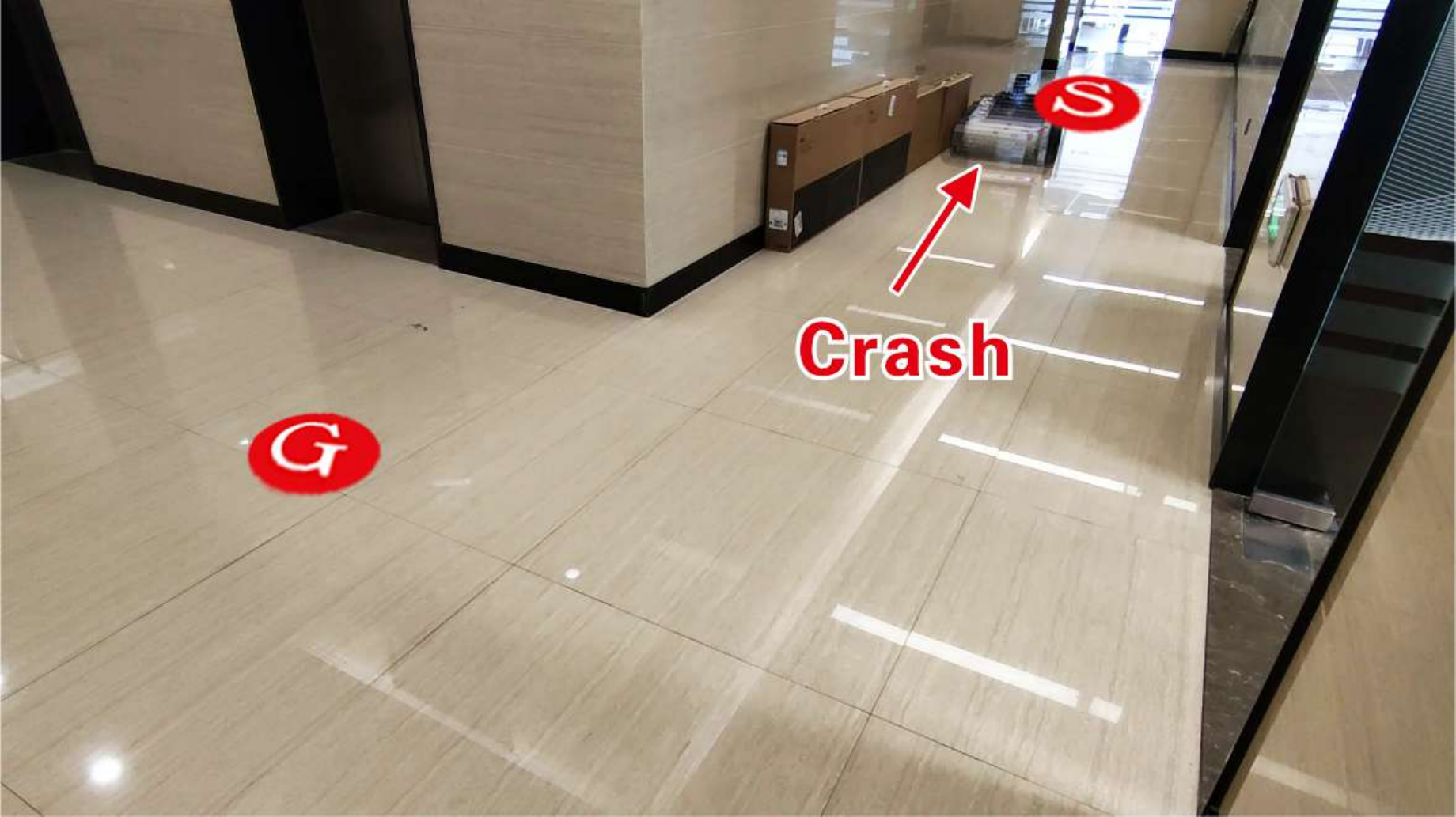}\label{dclp-r3}}

	\caption{Trajectories of robot trained with MAER-Nav and DRL-DCLP when tested in REnv1-3. The experimental videos can be found in the supplementary file.}
	\label{real_comparison_maer_dclp1}
\end{figure*}

\begin{figure}[t]
	\centering
    \subfloat[MAER-Nav in REnv4]{
\centering\includegraphics[width=0.485\linewidth]{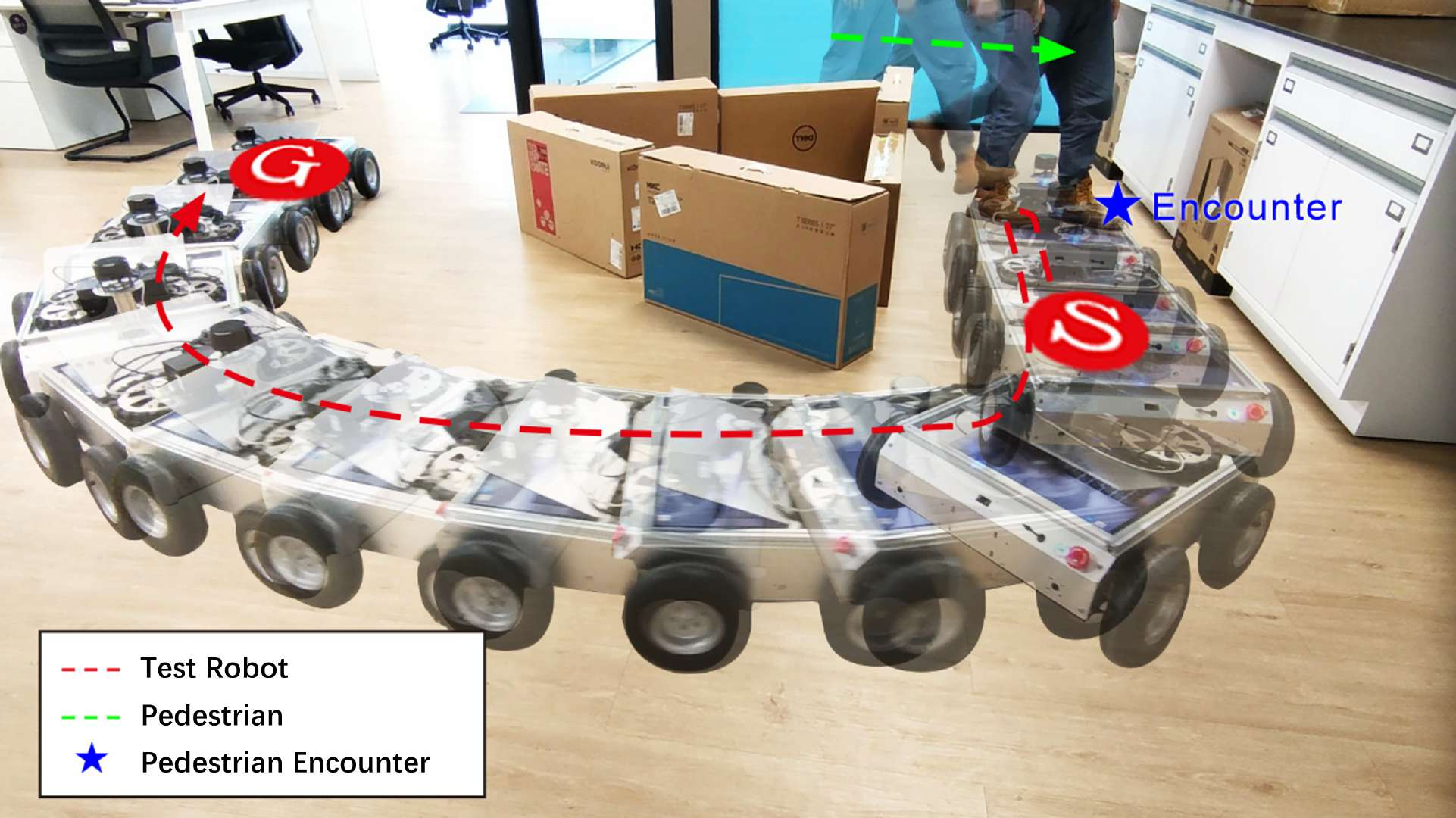}\label{maer-r4}}
    \hfill\subfloat[DRL-DCLP in REnv4]{
\centering\includegraphics[width=0.485\linewidth]{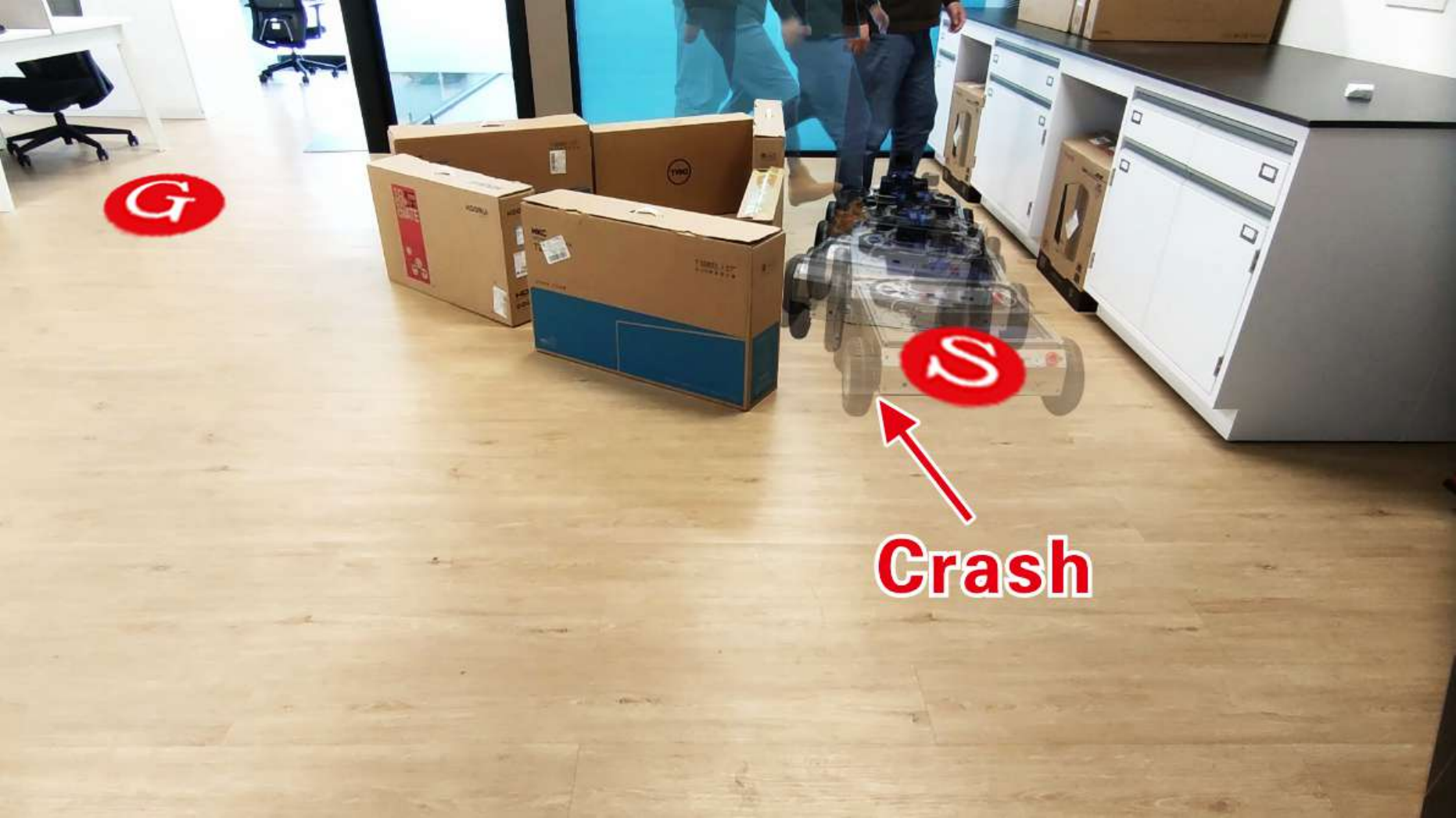}\label{dclp-r4}}
	
	\caption{Trajectories of robot trained with MAER-Nab and DRL-DCLP when tested in dynamic scenario REnv4. The experimental videos can be found in the supplementary file.}
	\label{real_comparison_maer_dclp2}
\end{figure}

\subsection{Hardware Setup}

As shown in Fig. \ref{robot}, the robot was built on the Wheeltec R550 Plus base and was fitted with a 360° 2D LiDAR LeiShen M10P, mounted at the front. The onboard local planner for the robot was a laptop, which has an i5-11320H processor. As MAER-Nav serves as the local planner, we utilized \textit{ROS GMapping} \cite{Gmapping} to create the map and \textit{ROS AMCL} \cite{AMCL} to precisely localize the robot within it. The map was used only for target localization, rather than for motion planning.

\subsection{Test Scenarios and Task Description}
We conducted experiments across four distinct real-world scenarios, as illustrated in Fig. \ref{realtestenv}. The first environment (REnv1) simulated a vehicle reversing out of a garage, representing a common situation where robots frequently encounter stucked situation. The second environment offered two configurations: REnv2-1 and REnv2-2, both featuring crowded spaces where the robot's expected initial movement direction was forward and backward, respectively. REnv3 presented a corridor scenario requiring backward navigation. Finally, REnv4 introduced dynamic pedestrian interactions to evaluate the robot's ability to execute backward recovery behaviors when forward movement became infeasible. In each test environment, the robot initiated its navigation from a position marked as ``S'' and was tasked with reaching the goal location designated as ``G''.

\subsection{Real-world Test Result Analysis}

The trajectories of both methods are presented in Fig.~\ref{real_comparison_maer_dclp1} and Fig.~\ref{real_comparison_maer_dclp2}, with corresponding videos available in the supplementary materials. Our proposed MAER-Nav method demonstrated exceptional navigation performance across all test environments, successfully completing all test tasks without collisions.
As shown in Fig.~\ref{maer-r2-1} and Fig.~\ref{dclp-r2-1}, we first confirmed that both methods exhibited comparable forward navigation capabilities. We then evaluated backward navigation capabilities across various scenarios to highlight the enhanced action flexibility of our method. Particularly in scenarios where traditional DRL-based methods typically encounter difficulties, as illustrated in Fig.~\ref{maer-r1} and Fig.~\ref{maer-r3}, our method successfully completed navigation tasks through effective backward maneuvers.
The dynamic pedestrian scenario provided a compelling demonstration of our method's adaptability. As depicted in Fig.~\ref{maer-r4}, the robot initially attempted forward movement but encountered pedestrian obstruction. While DRL-DCLP crashed due to its limited action flexibility (Fig.~\ref{dclp-r4}), MAER-Nav successfully implemented a reverse navigation strategy to escape the challenging situation. This clearly demonstrates that our method maintains the strong navigation performance of DRL-based approaches in complex environments while significantly enhancing action flexibility, enabling the robot to adapt to scenarios that would otherwise result in stuck state or collisions due to constrained motion capabilities.

\section{Conclusions}

This paper presented MAER-Nav, a novel DRL-based navigation framework that addresses the inherent action flexibility limitations in conventional DRL-based methods through a mirror-augmented experience replay mechanism. By transforming successful trajectories into synthetic experiences of opposite directionality, our approach enables the robot to learn effective bidirectional motion capabilities without requiring additional sensors or explicit reward modifications. Extensive evaluation in both simulation and real-world environments demonstrated that MAER-Nav achieves superior navigation performance compared to existing methods, with significantly improved success rates and reduced collision incidents across various challenging scenarios. The experimental results confirmed that our approach maintained the strong forward navigation capabilities of conventional DRL methods while substantially enhancing the robot's ability to execute backward maneuvers when necessary, particularly in confined spaces where traditional DRL-based approaches typically fail. MAER-Nav represents a substantial advancement in DRL-based mobile robot navigation, effectively bridging the gap between the comprehensive action space utilization of traditional planning methods and the complex environment handling capabilities of learning-based approaches.







\bibliographystyle{IEEEtran} 
\bibliography{mylib} 

\end{document}